%% file: main.tex
\def\year{2022}\relax
\documentclass[letterpaper]{article} 
\usepackage{aaai22}  
\usepackage{times}  
\usepackage{helvet}  
\usepackage{courier}  
\usepackage[hyphens]{url}  
\usepackage{graphicx} 
\usepackage{natbib}  
\usepackage{caption} 
\DeclareCaptionStyle{ruled}{labelfont=normalfont,labelsep=colon,strut=off} 
\usepackage{algorithm}
\usepackage{algorithmic}
\usepackage{comment}
\usepackage{subcaption}
\usepackage{newfloat}
\usepackage{listings}
\floatstyle{ruled}
\newfloat{listing}{tb}{lst}{}
\floatname{listing}{Listing}
\usepackage{xspace}
\newcommand{\scenic}[0]{\textsc{Scenic}\xspace}
\usepackage[dvipsnames]{xcolor}
\usepackage[T1]{fontenc}

\title{Programmatic Modeling and Generation of Real-Time Strategic Soccer Environments for Reinforcement Learning}
\author{
    Abdus Salam Azad\equalcontrib,
    Edward Kim\equalcontrib,
    Qiancheng Wu, Kimin Lee, \\ Ion Stoica, Pieter Abbeel, Alberto Sangiovanni-Vincentelli, Sanjit A. Seshia
}
\affiliations{University of California, Berkeley}

\begin{document}
\maketitle

\begin{abstract}
The capability of a reinforcement learning (RL) agent heavily depends on the diversity of the learning scenarios generated by the environment. Generation of diverse realistic scenarios is challenging for real-time strategy (RTS) environments. The RTS environments are characterized by intelligent entities/non-RL agents cooperating and competing with the RL agents with large state and action spaces over a long period of time, resulting in an infinite space of feasible, but not necessarily realistic, scenarios involving complex interaction among different RL and non-RL agents. Yet, most of the existing simulators rely on randomly generating the environments based on predefined settings/layouts and offer limited flexibility and control over the environment dynamics for researchers to generate diverse, realistic scenarios as per their demand. To address this issue, for the first time, we formally introduce the benefits of adopting an existing formal scenario specification language, \scenic{}, to assist researchers to \textit{model} and \textit{generate} diverse scenarios in an RTS environment in a flexible, systematic, and programmatic manner. To showcase the benefits, we interfaced \scenic to an existing RTS environment Google Research Football(GRF) simulator and introduced a benchmark consisting of 32 realistic scenarios, encoded in \scenic, to train RL agents and testing their generalization capabilities. We also show how researchers/RL practitioners can incorporate their domain knowledge to expedite the training process by intuitively modeling stochastic programmatic policies with \scenic.


\end{abstract}

\section*{Introduction}
\input{sections/intro}

\input{sections/relatedwork}
\input{sections/background}
\input{sections/scenic}
\input{sections/evaluation}

\input{sections/conclusion_futurework}
\input{sections/acknowledgment}

\bibliography{main}
\newpage
\appendix
\input{sections/appendix}

\end{document}


\maketitle

\input{Supplimentary Materials/supplementary}

\bibliographystyle{ref_bst}
\bibliography{reference}


%% file: sections/intro.tex
Deep reinforcement learning (RL) has emerged as a powerful method to solve a variety of sequential decision-making problems, including board games~\cite{silver2017mastering,silver2018general}, video games~\cite{mnih2015human,vinyals2019grandmaster}, and robotic manipulation \cite{kalashnikov2018qt}. These successes rely heavily on widely-used simulation environments~\cite{bellemare2013arcade, brockman2016openai} and benchmarks~\cite{ cobbe2020leveraging, duan2016benchmarking,dmcontrol_old}. However, regardless of a long history of RL benchmarks, the existing RL environments/simulators are insufficient to properly train, test, and benchmark RL algorithms for real-time strategy (RTS) environments such as Starcraft~\cite{starcraft}, Dota2~\cite{openai2019dota}, and soccer~\cite{kurach2020google}, due to their lack of support for modeling diverse scenarios involving sophisticated interactive behaviors.

These RTS environments are characterized by unique characteristics that require special support for modeling. The environments involve intelligent entities/non-RL agents co-operating and competing with the RL agents with large state and action spaces over a long horizon. This opens up extremely diverse strategies consisting of numerous interactive behaviors. Yet, most of the existing simulators rely on randomly generating the environments based on predefined settings/layouts and offer limited flexibility and control to the researchers over the environment dynamics to generate diverse realistic scenarios. As a result RL research face at least two fundamental challenges: (i) the lack of diverse and realistic training data often leads to lack of generalization~\cite{cobbe2019quantifying, cobbe2020leveraging,lee2019network,lee2020context}, and (ii) the lack of flexibility and control over the environment dynamics makes it hard to generate realistic evaluation scenarios to comprehensively test generalization in these complex RTS environments. 



To address this issue, for the first time to the best of our knowledge, we introduce the benefits of adopting an existing formal scenario specification language, \scenic{}, to assist researchers to \textit{model} and \textit{generate} diverse realistic scenarios in an RTS environment in a flexible, systematic, and programmatic manner. Each \scenic{} program represents a Markov Decision Process (MDP) and provides high-level syntax and semantics, backed by its own compiler, to intuitively and quickly model diverse and complex interactive scenarios to train RL agents and test their generalization capabilities. Furthermore, it allows researchers/RL practitioners to incorporate their domain knowledge into the training process by generating offline data with stochastic programmatic policies written in high-level intuitive syntax of \scenic{}. To demonstrate the benefits, we interfaced \scenic{} to an existing RTS environment, Google Research Football (GRF)~\cite{kurach2020google}.


\newpage
Our contributions are as follows:
\begin{itemize}
    \item For the first time, we introduce the benefits of adopting a scenario specification language to (1) flexibly model interactive scenarios to train RL agents, (2) test their generalization capability, and (3) program stochastic RL policies to generate demonstration data.
    
    \item We open-sourced our \scenic{}'s interface to GRF environment along with our 32 scenarios, 5 stochastic policies, and libraries encoded in \scenic{} to assist researchers to build upon them to easily model diverse and sophisticated scenarios.
    
\end{itemize}

    
    
        
        
    

\begin{figure*}[t!]
    \centering
    \begin{subfigure}[t]{0.2\textwidth}
        \centering
        \includegraphics[height=0.95in]{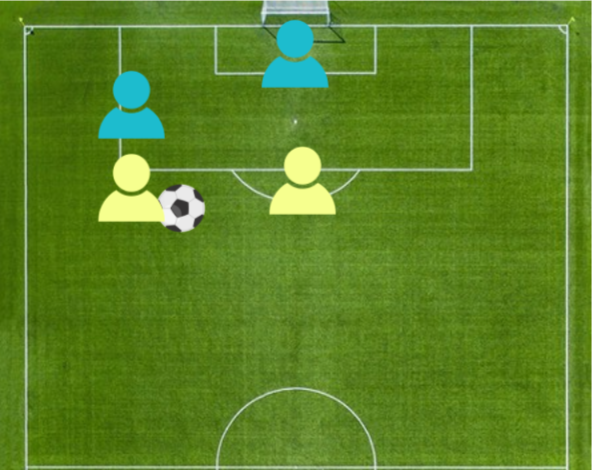}
        \caption{a bird-eye view of the scenario}
        \label{fig:pass_and_shoot_birdeye}
    \end{subfigure}
    ~
    \begin{subfigure}[t]{0.2\textwidth}
        \centering
        \includegraphics[width=1.2in,height=0.95in]{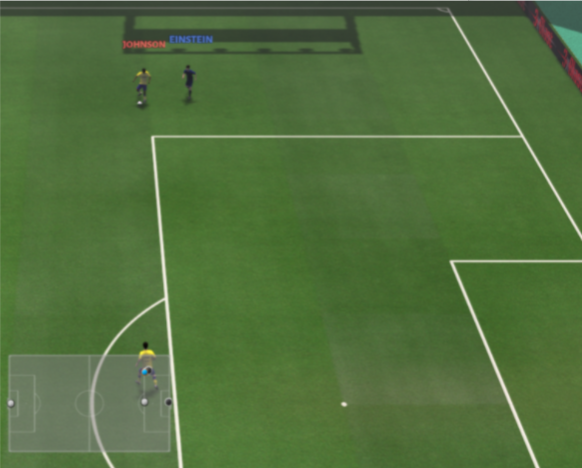}
        \caption{a snapshot of GRF environment}
        \label{fig:grf_snapshot}
    \end{subfigure}
    ~
    \begin{subfigure}[t]{0.45\textwidth}
        \centering
        \includegraphics[width=\linewidth]{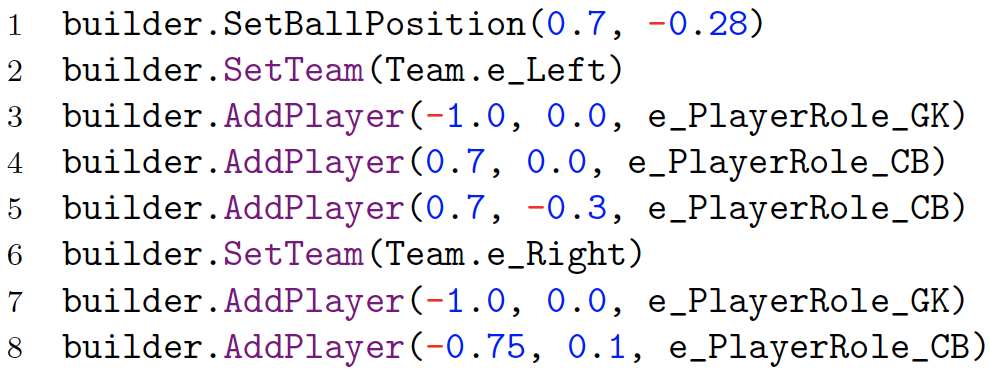}
        \caption{GRF's scenario program}
        \label{fig:grf_scenario}
    \end{subfigure}
    ~
    \begin{subfigure}[t]{0.7\textwidth}
        \centering
        \includegraphics[width=\linewidth]{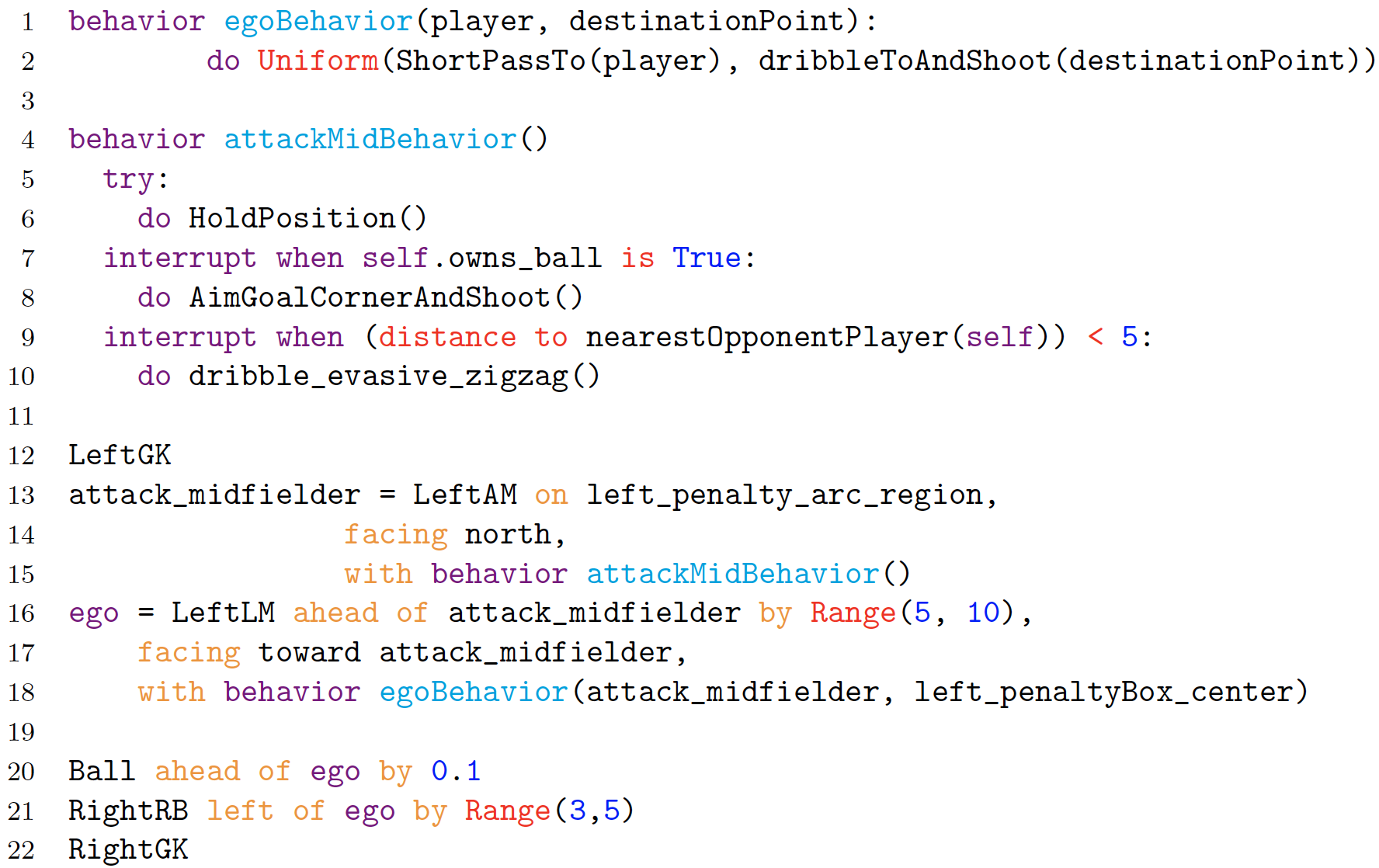}
        \caption{\scenic{} program of generalized pass-and-shoot scenario with distribution over players' initial condition and behaviors}
        \label{fig:scenic_program}
    \end{subfigure}
    \caption{Programs encoding the Google Research Football's (GRF) pass-and-shoot scenario}
    \label{fig:pass_and_shoot}
\end{figure*}

%% file: sections/relatedwork.tex
\section*{Related Work}

\textbf{Environment Generation in RL}: In literature, several techniques have been adopted to generate a rich variation of learning scenarios, primarily to promote or, ensure generalization. Techniques such as changing background with natural videos~\cite{zhang2018natural}, introducing sticky actions~\cite{machado2018revisiting} have been attempted, but are not robust enough. To ensure generalization, \cite{lee2020context} and \cite{seo2020trajectory} generated training and testing scenarios by randomly sampling from different regions of parameter space. Similar to supervised learning, the use of separate train and test sets have also been adopted~\cite{nichol2018gotta, cobbe2020leveraging, cobbe2019quantifying, justesen2018illuminating}, typically using techniques such as Procedural Content Generation~\cite{procedural}, which has traditionally been used to automatically generate levels in video games. However, most of these focus on discrete domain, typically the dataset generation process is opaque, and it can be difficult to  quantify or, reason about how different (or, similar) these train and test sets are, because the generation process often use random numbers to generate different configurations.

On the contrary, a few manually scripted scenario benchmarks are proposed with respect to a few RTS RL environments with limitations. For StarCraft~\cite{starcraft}, only two benchmark scenarios~\cite{starcraft_scenario1, starcraft_scenario2} have been proposed. Both of these model different initial states but leave the behavior generation to either a learned RL agent or AI bots that are provided by the StarCraft environment, which are considered as blackbox agents. As a result, a sophisticated modeling and control over the behaviors of non-RL agents to create specific types of scenarios is not possible, severely restricting the diversity of the scenarios. For soccer domain, \citet{soccer_scenario2} presented one benchmark scenario on keepaway tactical scenario and later extended to more general half-field offense scenario~\cite{soccer_scenario1} and provided a library of APIs relating to behaviors (e.g. mark player, defend goal) of players, which helps users to model scenarios. However, \scenic{} provides further benefits that are not covered in this work. \scenic{} provides high-level syntax and semantics to (i) easily write spatial relations for intutively modeling initial states, (ii) assign distributions over both initial states and behaviors to generate variations of environments for robust training and testing generalization, and (iii) specify priorities over interaction conditions over behaviors to model more sophisticated types of higher level behavior (for more detail refer to Section \textit{Modeling Scenarios with \scenic{}}).

\textbf{Formal Scenario Specification Languages for Environment Modeling and Generation} A few scenario specification languages have been proposed in the autonomous driving domain including \scenic{}. Paracosm language ~\cite{paracosm} models dynamic scenarios with reactive and synchronous model of computation. The Measurable Scenario Description Language (M-SDL)~\cite{msdl} shares common features as \scenic{} to model interactive scenarios. In contrast, however, \scenic{} provides a much higher-level, probabilistic, declarative way of modeling. Furthermore, unlike other scenario specification languages, \scenic{} has demonstrated its generality over different domains such as autonomous driving, robotics, and aviation~\cite{scenic-dynamic}. For these reasons, we chose \scenic{} in this paper for demonstration of benefits that a scenario specification language can provide to RL.

%% file: sections/background.tex
\section*{Background}

\subsection*{Google Research Football Simulator}\label{sec:grf_background}
The Google Research Football (GRF) simulator~\cite{kurach2020google} provides a realistic soccer environment to train and test RL agents. The setting, the rules, and the objective of the environment are the same as defined by Fédération Internationale de Football Association~\cite{fifa}. The environment setup is as the following. All the players on the field are controlled by (1) GRF's built-in, rule-based AI bots and (2) RL agents. The simulator dynamically determines which of the RL team players are to be controlled by RL agents based on their vicinity to the ball. GRF provides 11 offense scenarios to train and test RL agent performance and it provides trained RL agent checkpoints for a subset of its scenarios. 

\subsection*{Scenario Specification Language: \scenic{}}\label{sec:scenic_background}
\scenic{}~\cite{scenic-static,scenic-dynamic} is an object-oriented, probabilistic programming language whose syntax and semantics are designed to intuitively \textit{model} and \textit{generate} scenarios. A \scenic{} program represents an abstract scenario, which models a \textit{distribution} over initial states and behaviors of players in the scenario. For each scenario generation, an initial state is sampled from the program at the beginning of a simulation and interactive behaviors are sampled during simulation runtime. Therefore, with a single \scenic{} program, users can generate a distribution of concrete scenarios. 

\scenic{} requires action and model libraries, which are imported and compiled with a user's \scenic{} program for execution. The action library defines the action space which is determined by the simulator. The model library defines objects and their attributes (e.g. position, heading). We can assign prior distributions over these attributes. For example, a goalkeeper's position can be uniformly randomly distributed over the penalty box region. If a user simply instantiates a goalkeeper in a \scenic{} program but does not specify any condition over its attributes, then they are sampled from the prior distributions by default. These prior distributions can be overwritten in the user's \scenic{} program.


%% file: sections/scenic.tex
\section*{Scenario Specification Language for RL}\label{scenic_for_RL}

\subsection*{Benefits of Scenario Specification Language for RL}
The objective of this paper is to introduce the benefits of the use of scenario specification language for modeling and generating scenarios, specifically for RTS environments for RL. Using a scenario specification language whose syntax and semantics are carefully designed to intuitively model scenarios have the following benefits:

\begin{enumerate}
    \item \textbf{Easily Model Interactive Environments on \textit{User-demand} to Train and Test RL Agents:} The intuitive syntax and semantics, which abstracts away the implementation details and allows users to reason solely at high-level semantics, makes it easy to model complex spatial relations among multiple agents, their behaviors and conditions on how these behaviors should interact. It should be noted that, it requires a considerable amount of research and engineering effort to design and implement a formal scenario modeling language and its compiler from scratch.
    \item \textbf{Program Stochastic Policies:} These programmed agents can serve two purposes: (i) allow developers to incorporate domain knowledge, e.g., generate demonstration data for offline training and (ii) provide performance baseline for trained RL agents.
    \item \textbf{Interpretability and Transparency}: 
    The intuitive syntax and semantics make scenario programs interpretable and transparent. Therefore, users can reason about the difference/similarity of train and test environments by comparing their scenario programs.
    \item \textbf{Reusability of Existing Scenarios}: The interpretability of scenario programs facilitates easy modification or re-use of existing \scenic\ programs, models, and behaviors to quickly model new scenarios. This facilitates building a community around designing and sharing scenario programs, by building upon each other's scenarios.
\end{enumerate}

\begin{figure*}[t!]
    \begin{subfigure}[t]{0.23\textwidth}
        \centering
        \includegraphics[height=1.6in, width=1.4in]{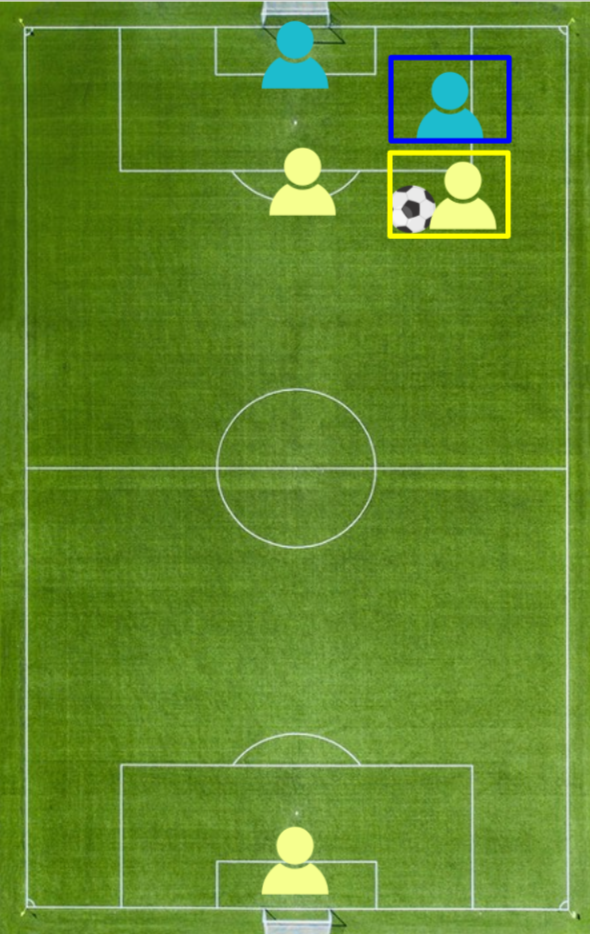}
        \caption{generalization test scenario for the scenario in Fig.~\ref{fig:pass_and_shoot_birdeye}}
        \label{fig:test_pass_and_shoot}
    \end{subfigure}
    ~
    \begin{subfigure}[t]{0.23\textwidth}
        \centering
        \includegraphics[height=1.6in, width=1.4in]{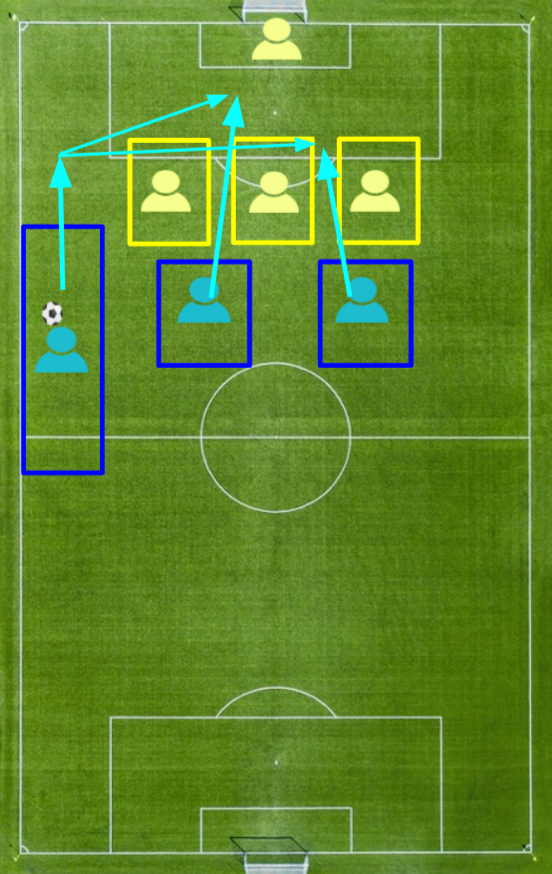}
        \caption{3 vs 3 left mid-fielder crosses to either player in penalty box}
        \label{fig:defense_3vs3}
    \end{subfigure}
    ~
    \begin{subfigure}[t]{0.23\textwidth}
        \centering
        \includegraphics[height=1.6in, width=1.4in]{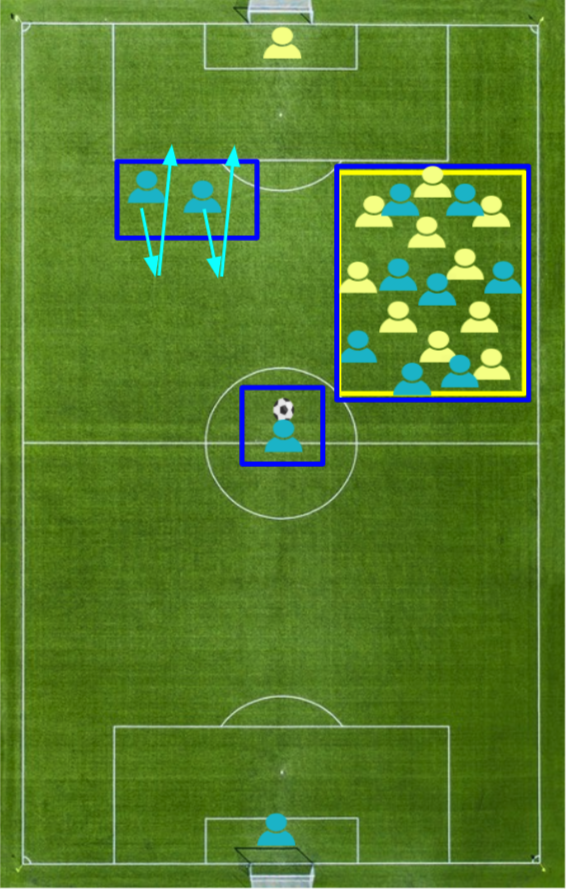}
        \caption{11 vs 11 open player scenario}
        \label{fig:vision}
    \end{subfigure}
    ~
    \begin{subfigure}[t]{0.23\textwidth}
        \centering
        \includegraphics[height=1.6in, width=1.4in]{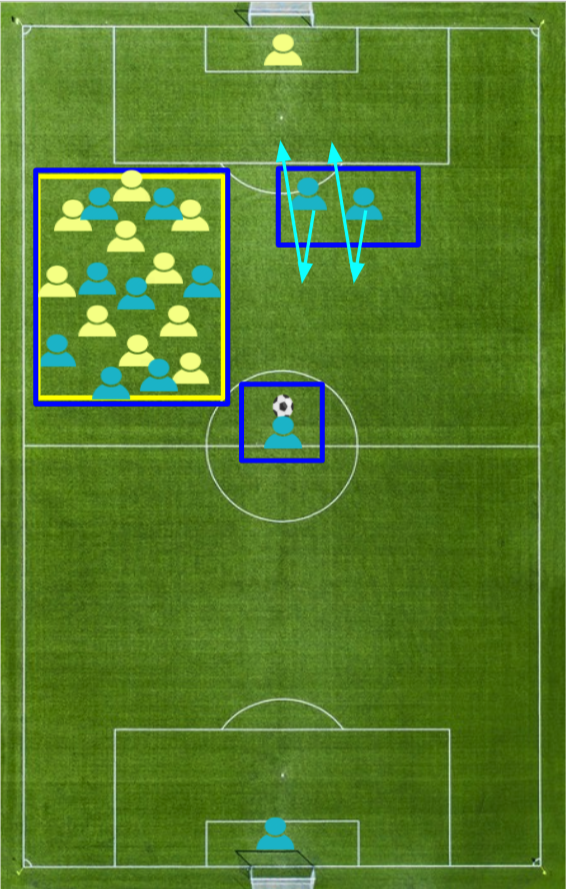}
        \caption{mirrored Fig.~\ref{fig:vision} scenario}
        \label{fig:vision_mirrored}
    \end{subfigure}
    
    \caption{Examples of a new defense scenarios with specific assigned behaviors (a), a test scenario to assess generalization (b), and two full game scenarios (c,d) we used for training and testing. The RL team is yellow and the opponent, blue. The assigned opponent behaviors are highlighted with light blue arrows. Uniformly random distribution is assigned over a specific region for each player. These regions are highlighted boxes.}
    \label{fig:new_scenarios}
\end{figure*}

\begin{figure}
    \centering
    \includegraphics[width=\linewidth]{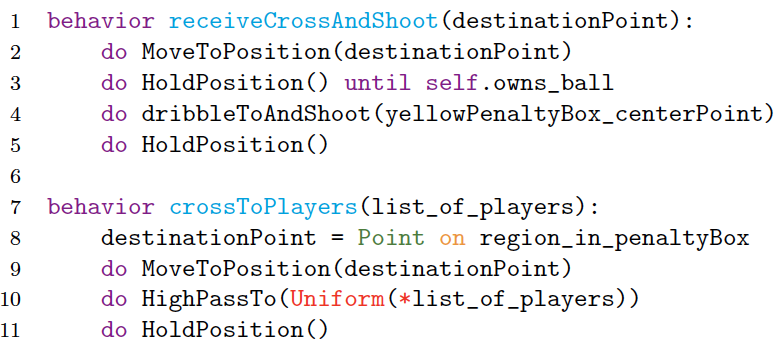}
    \caption{A snippet of a \scenic{} program specifying behaviors for players Fig.~\ref{fig:defense_3vs3}}
    \label{fig:example_behaviors}
\end{figure}

\newpage
\subsection*{Modeling Scenarios with \scenic{}}
Formally, a scenario is a Markov Decision Process (MDPs)~\cite{sutton2018reinforcement} defined as a tuple $\left( \mathcal{S}, \mathcal{A}, p, r, \rho_0\right)$, with $\mathcal{S}$ denoting the state space, $\mathcal{A}$ the action space, $p\left(s^\prime|s,a\right)$ the transition dynamic, $r\left(s,a\right)$ the reward function, and $\rho_0$ the initial state distribution. Given the state and action spaces as defined by the GRF environment, a \scenic{} program defines (i) the initial state distribution, (ii) the transition dynamics (specifically players' behaviors), and (iii) the reward function. Hence, users can exercise extensive control over the environment with \scenic{}.

{\bf Modeling Initial State Distribution} 
Users can intuitively specify initial state distributions with \scenic{}'s high-level syntax that resembles natural English. For example, refer to the full \scenic{} program in Fig.~\ref{fig:pass_and_shoot}(d) which describes a more generalized version of GRF's Pass and Shoot scenario as visualized in Fig.~\ref{fig:pass_and_shoot}(a,b). In line 12-22, the initial state distribution is specified. The \scenic{} syntax for modeling spatial relations among players are highlighted in yellow. In addition, \scenic{} supports about 20 different syntax to support modeling complex spatial relations~\cite{scenic-dynamic}. Rather than having to hand-code positions for a concrete scenario as in the GRF's scenario~\ref{fig:pass_and_shoot}(c), users can much more intuitively and concisely model a distribution of initial states. Here, \texttt{Left} represents the yellow team, \texttt{Right} the blue, and the two following abbreviated capital letters indicate the player role. 


{\bf Modeling Transition Dynamics}
One can flexibly modify transition dynamics of the environment by specifying the behaviors of non-RL players using \scenic{}. Take the same example \scenic{} program in Fig.~\ref{fig:pass_and_shoot}(d) as above. Line 1-10 models two new behaviors. A behavior can invoke another behavior(s) with syntax \texttt{do}, succinctly modeling a behavior in a hierarchical manner. Users can assign distribution over behaviors as in line 2. The interactive conditions are specified using try/interrupt block as in line 5-10. Semantically, the behavior specified in the try block is executed by default. However, if any interrupt condition is satisfied, then the default behavior is paused and the behavior in the interrupt block is executed until completion and then the default behavior resumes. These interrupts can be nested with interrupt below has higher priority. In such case, the same semantics is consistently applied. 

{\bf Rewards}
\scenic{} has a construct called \texttt{monitor}, which can be used to specify reward functions. The reward conditions in the \texttt{monitor} is checked at every simulation step and updates the reward accordingly. 

{\bf Termination Conditions} 
Users can also specify termination conditions which are monitored at every simulation time step. 

\subsection*{On Interfacing \scenic{} to a Simulator}
Interfacing \scenic{} to other simulators is straight-forward. In fact, \scenic{} is already interfaced with five other simulators~\cite{scenic_supported_sim} in domains such as autonomous driving, aviation, and robotics. To interface \scenic{}  with a simulator, one needs define the model, action, and behavior libraries. These libraries expedites modeling complex scenarios by helping users re-use the set of models, actions, and behaviors in the libraries, rather than having to write a scenario from scratch. 

The \textit{model} library defines the state space. It defines players with distribution over their initial state according to their roles and GRF's AI bot is assigned by default to all player behavior. These prior distribution over the initial state and behavior can be overwritten in the \scenic{} program. The model library also defines region objects such as goal and penalty box regions as well as directional objects in compass directions. The \textit{action} library defines the action space as determined by the GRF simulator. These action space consists of movement actions in eight compass directions, long/short/high pass, shoot, slide, dribble, and sprint. 

The \textit{behavior} library consists of behaviors and helper functions that represent widely used basic skills in soccer. These behaviors include give-and-go, evasive zigzag dribble to avoid an opponent's  ball interception, dribbling to a designated point and shooting, shooting towards the left or right corner of the goal, etc. Additionally, the behavior library also include useful helper functions such as identifying nearest opponent or teammate, whether there is an opponent near the running direction of a dribbler, etc. Please refer to our open-sourced repository for more details.

\begin{figure}[t!]
  \includegraphics[scale=0.4]{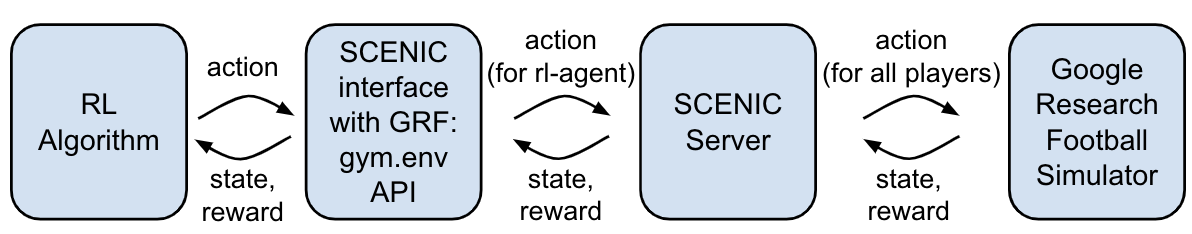}
  \caption{Interface Architecture between \scenic{} and GRF}
  \label{fig:rlarch}
\end{figure}

\begin{figure*}[t!]
  \centering
  \includegraphics[scale=0.3]{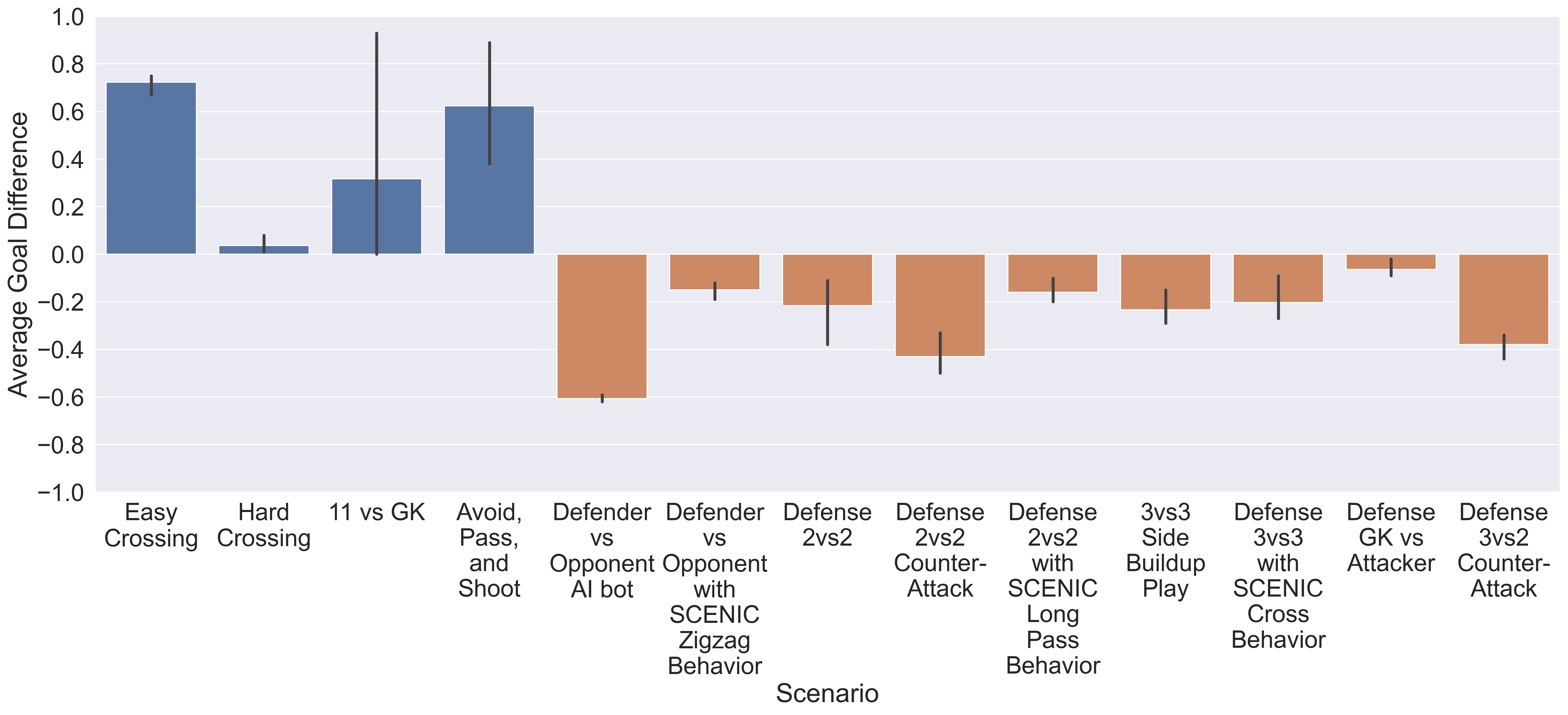}
  \caption{Average Goal Difference of PPO agents on the proposed mini-game scenario benchmark. The error bars represent 95\% bootstrapped confidence intervals}
   \label{fig:scenario-training}
\end{figure*}

\subsubsection{Interface Architecture}
Figure~\ref{fig:rlarch} shows an overview of our overall architecture. The architecture can be divided into two parts: i) RL interface, through which the RL algorithms interact with \scenic \ and ii) the \scenic\ Server, which executes a \scenic{} program and governs the simulation by interacting with the underlying simulator. We follow the widely used OpenAI Gym API~\cite{brockman2016openai} as our interface, which allows our interface to be used seamlessly with all the existing standard RL frameworks.

For each simulation/episode, the \scenic\ server first samples an initial state from the \scenic{} program to start a new scenario in the GRF simulator and updates its internal model of the world (e.g., player and ball positions). From then on, a round of communication occur between the RL algorithm and \scenic\ server, with the RL interface at the middle. At each timestep, the gym interface takes in the action(s) for the RL agent and passes them to the \scenic\ server. The \scenic\ server in turn computes actions for all the remaining non-RL players---the players not controlled by the RL agent---and then executes all these actions (of both the RL and non-RL players) in the simulator. The \scenic\ server then receives the observation and reward from the simulator, updates the internal world state, and then passes them back to the RL algorithm. This interaction goes on till any terminating conditions as specified in the scenario script is satisfied.


%% file: sections/evaluation.tex
\section*{Evaluation}\label{sec:evaluation}
In this section, we demonstrate four use cases of \scenic{} in RL. First, we present and benchmark a set of 13 realistic mini-game scenarios encoded in \scenic{} with a varying level of difficulty. Second, we test the generalization capabilities of the trained RL agents on unseen, yet intuitively similar scenarios. Next, we show how developers can ``debug" their agents for failure scenarios of interest. At last, we show how probabilistic \scenic\ policies can be used to generate offline data and endow domain knowledge into the learning process for faster training, which we believe to be very important for applying RL in practice.


\subsection*{Experimental Setup}\label{sec:exp_setup}
We run PPO~\citep{ppo} on a single GPU machine (NVIDIA T4) with 16 parallel workers on Amazon AWS. Unless otherwise specified, all the PPO training are run for 5M timesteps and repeated for 10 different seeds. All the evaluation has been done for 10000 timesteps. For all the experiments, we use the stacked Super Mini Map representation for observations ---a 4x72x96 binary matrix representing positions of players from both team, the ball, and the active player---and the scores as rewards, i.e., $+1$ when scoring a goal and $-1$ upon conceding, from~\cite{kurach2020google}. Similar to the academy scenarios from~\cite{kurach2020google}, we also terminate a game when one of the following happens: either of the team scores, ball goes out of the field, or, the ball possession changes.  For further details, including hyperparameters and network architecture, we refer readers to the Supplementary Materials (Section \textit{Details on Experimental Setup and Training}).

\subsection*{Mini-game Scenario Benchmark}
Training an RL agent to solve a full soccer game involving 22 players is very challenging and may take days even with distributed algorithms. For example, ~\citet{kurach2020google} showed even the easy version of GRF's 11 vs 11 game cannot be solved with 50M samples. To allow researchers to iterate their ideas with a reasonable amount of time and compute, we present a set of 13 mini-game scenarios. All these scenarios are inspired from common situations occurring in real soccer games but involves fewer number of players to make them amenable to be faster training. 

Nine of our proposed mini-game scenarios are defense scenarios, which are nice complement to GRF's offense-only scenarios (refer to Sec.~\ref{sec:grf_background}), along with four new offense scenarios. Most of these scenarios are initialized from a distribution, rather than fixed locations. By default all the opponent players are controlled by GRF's built-in AI bot (refer to Sec.~\ref{sec:grf_background}). However, for the scenarios where the AI bot does not exhibit our desired behavior, we model the opponent behaviors using \scenic. For example, in the 3vs3 cross scenario as shown in Fig.~\ref{fig:defense_3vs3}, the opponent AI bots tried to pass the ball around instead of crossing. Therefore, we modelled and assigned behaviors such that the blue player on the leftmost side of the field would run up the field and cross the ball. Meanwhile, the two blue players in the center run into the penalty box area to receive the cross and shoot. These modelled behaviors are shown in Fig.~\ref{fig:example_behaviors}. 

We benchmark our mini-game scenarios by training agents with PPO. Figure \ref{fig:scenario-training} shows the average goal differences for all the scenarios. For these mini-game scenarios, we end the game if one of the teams score. Hence, the goal difference can range between -1 to +1. For the offense scenarios, a well trained agent is supposed to score consistently achieving an average goal difference close to +1. On the other hand, a well-trained agent should achieve a goal difference close to 0 for successfully defending the opponents in the defense scenarios. From the graph it can be seen that the proposed scenarios offer a varied levels of difficulties. For example, PPO consistently achieves goal difference of around 0.5 for the {\sc Easy Crossing} scenario, but barely learns anything for {\sc Hard Crossing}. In case of the defense scenarios, the results also show a varied range of difficulty, {\sc GK vs Opponent} scenario being be easiest. 




\subsection*{Testing for Generalization}

\begin{figure*}[ht!]
  \begin{subfigure}[t]{\textwidth}
  \centering
  \includegraphics[scale=0.35]{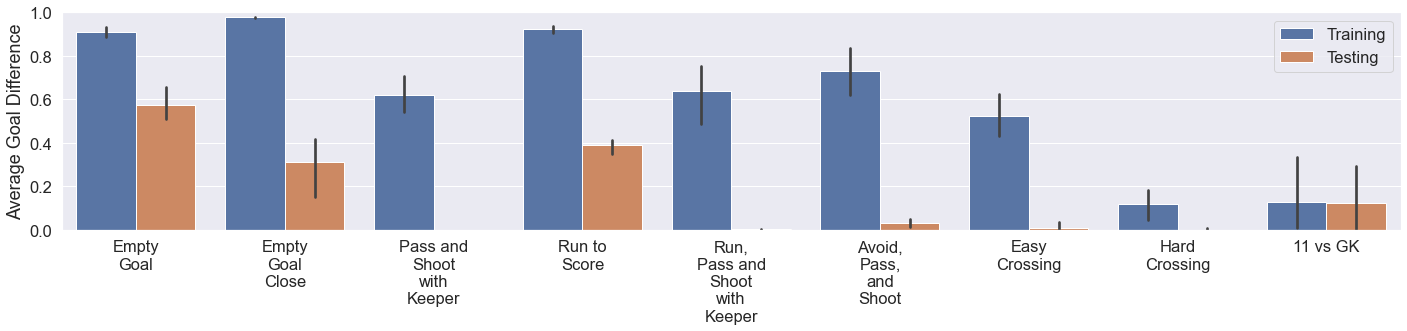}
  \caption{Offense and select GRF academy scenarios}
   \label{fig:testing_offense}      
  \end{subfigure}
 
  \begin{subfigure}[t]{\textwidth}
  \centering
  \includegraphics[scale=0.35]{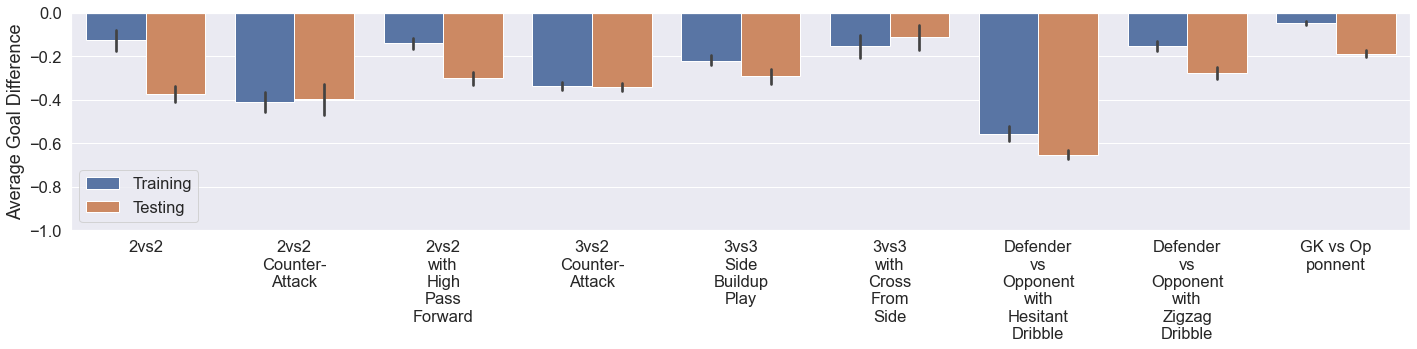}
  \caption{Defense scenarios}
   \label{fig:testing_defense}      
  \end{subfigure}
  \caption{Evaluation of PPO agents' generalization against varying initial conditions. For most of the academy and offense scenarios we observe a significant drop in performance. However, for several defense scenarios the difference in train and test scenarios is not that significant.}
   \label{fig:testing_for_gen}
\end{figure*}

We provide scripts to test generalization of all of our 13 new benchmark scenarios along with 5 scenarios provided by GRF. We changed the distribution over the initial state while keeping the formation of players and their behaviors in each scenario intact. For example, for testing generalization of an RL agent trained in the Pass and Shoot scenario (Fig.~\ref{fig:pass_and_shoot_birdeye}), we instantiated the yellow and the blue players on the symmetric right side of the field instead of the left and kept the other initial state distribution the same (Figure~\ref{fig:test_pass_and_shoot}). 

Fig.~\ref{fig:testing_for_gen} compares the trained agents' performance in training and test scenarios. As expected, we observe a noticeable drop of performance in most of the GRF's academy and offense scenarios (Fig.~\ref{fig:testing_offense}). For example, the Pass and Shoot scenario (Figure~\ref{fig:pass_and_shoot_birdeye}), which achieved around 0.6 in training, failed to generalize for the test scenario. However, for the defense scenarios, the drop in performance was not as noticeable. We conjecture that this distinction comes from the differences in the offense and defense training scenarios, where the defense scenarios tend to contain larger distribution over the initial state than those of the offense scenarios (refer to Supplement). Consequently, larger variations of scenarios introduced during training may have contributed to better generalization for defense scenarios. 

\begin{figure*}[h!]
  \centering
  \includegraphics[scale=0.35]{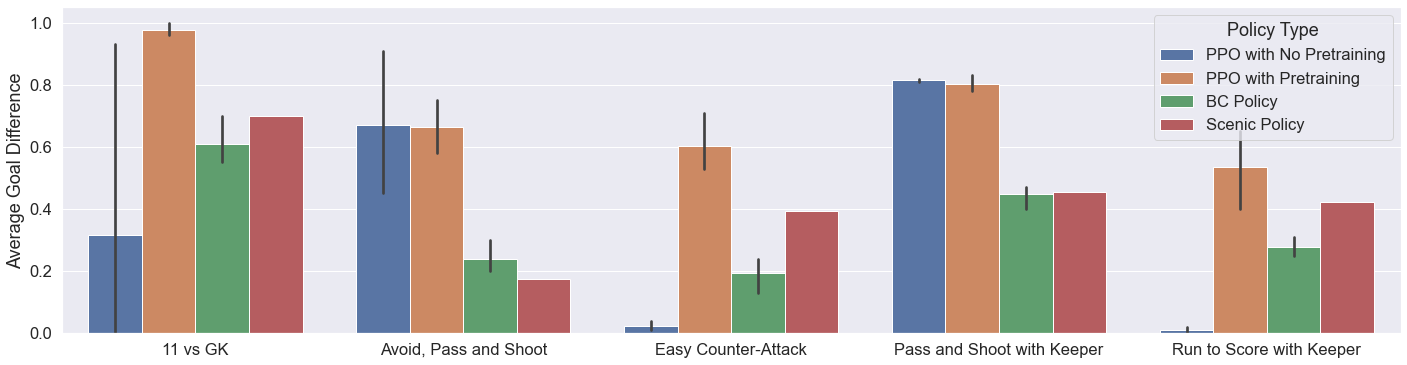}
  \caption{Performance of PPO agents trained with and without any demonstration data, along with the performance of corresponding behavior-cloned and \scenic\ policies. We see significantly better performance on three of the scenarios, while the rest two achieves comparable performance, highlighting the usefulness of the proposed \scenic\ policies.}
   \label{fig:pretrain}
\end{figure*}
\subsection*{Debugging Agents on 11v11 Failure Scenario}
For this experiment, we evaluate and debug an RL checkpoint provided by GRF, which was trained on their 11 vs 11 easy stochastic scenario, i.e., easy version of their full-game scenario. This agent achieves an impressive average goal difference of 6.99 per full-game\footnote{Evaluated on 100K timesteps}, scoring up to 14 goals in the training scenario during our experiments. We modelled a scenario, as visualized in Figure~\ref{fig:vision}, to test the agent's ability to quickly perceive open teammates near the opponent goal to advance the ball forward and score---a crucial skill for soccer. When we assigned GRF's built-in AI bots to control the open players on the left side of the field, the players ran straight toward the ball, instead of taking advantage of the closeness to the opponent goal without being marked. Hence, we modelled a behavior for open players in \scenic{} so that they would stay close to the goal while abiding by the offside rule.

Although obvious to humans, the trained checkpoint performs poorly in this scenario with an average goal difference of 0.1. To `debug' the agent, we then fine-tune the agent on a `mirrored' scenario, as shown in Figure~\ref{fig:vision_mirrored}, with PPO for 5M timesteps. The fine-tuned agent improved noticeably on the original scenario, achieving an average goal difference of 0.67. This showcases the usefulness of \scenic{} to easily model and generate scenarios of interest using one's domain knowledge, which may have been difficult with blackbox agents (e.g. built-in AI bots, or trained RL agents), to test and debug certain capabilities of an RL agent. 

     


\subsection*{Facilitating Training with Probabilistic \scenic\ Policies}

In the section, we show how RL practitioners can incorporate their domain knowledge by writing probabilistic \scenic\ policies for faster training. We wrote simple semi-expert RL policies for five different scenarios, where the agent suffers to learn, and generated 8K samples of demonstration data for per scenario. To facilitate training on those scenarios, we first pre-train an agent via behavior cloning with the generated offline data and then fine-tune the agent using PPO for 5M timesteps. All the experiments were repeated for three different seeds. Figure ~\ref{fig:pretrain} compares the training performance of these agents against the agents that were trained with PPO only. We notice that, even with such a low volume of demonstration data, we can train much better agents and can solve scenarios which were otherwise unsolved. The experimental results thus suggests, with stochastic \scenic{} policies we can generate rich quality demonstration data to substantially enhance training performance, which can be particularly useful in practice for environments like GRF which requires a heavy compute resource.

%% file: sections/conclusion_futurework.tex
\section*{Conclusion \& Future Work}\label{sec:conclusion}
We introduced and demonstrated the benefits of adopting a scenario specification language to train RL agents and test their generalization capabilities in various realistic scenarios generated by \scenic{} programs, which succinctly capture distributions of initial states and behaviors. We also showcased modeling domain knowledge via stochastic \scenic{} policies by generating demonstration data to facilitate training in GRF, a complex real-time strategy environment. We hope our work could gather an interest to support systematic modeling of RTS environments. 

%% file: sections/acknowledgment.tex
\section*{Acknowledgements}
This work was supported in part by National Science Foundation grants CNS-1730628,  CNS-1545126 (VeHICaL), CNS-1739816, and CCF-1837132, by DARPA contracts FA8750-16-C0043 (Assured Autonomy) and FA8750-20-C-0156 (Symbiotic Design of Cyber-Physical Systems), by Berkeley Deep Drive, by Toyota through the iCyPhy center, by the Toyota Research Institute, and by the Berkeley Artificial Intelligence Research (BAIR) Commons program. 

%% file: sections/appendix.tex
\section{Description of Proposed Scenarios and Policies}\label{sec:dataset_descrition}
In this section we provide brief descriptions of all of the scenarios in our dataset. To see our \scenic{} programs, please refer to our attached README pdf file for the pathways to our scenarios. 

\subsection{On Mini and Full Game Scenarios}
In general, all our scenarios have the following three termination conditions: (i) ball goes off the field, (ii) change in ball possession across teams, (iii) one of the team scores. If any of these conditions are satisfied, then the scenario will terminate in simulation. 

\subsubsection{Offense Scenarios} 

\begin{figure*}
     \centering
     \begin{subfigure}[b]{0.45\textwidth}
         \centering
         \includegraphics[width=\textwidth]{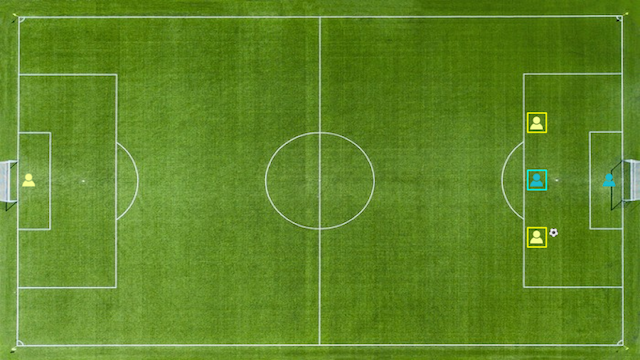}
         \caption{{\sc Easy Crossing}}
     \end{subfigure}
     \hfill
     \begin{subfigure}[b]{0.45\textwidth}
         \centering
         \includegraphics[width=\textwidth]{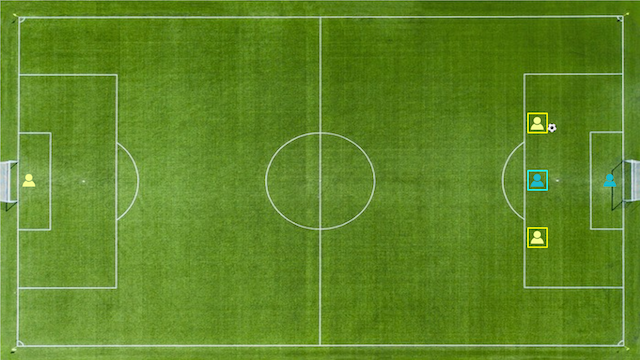}
         \caption{{\sc Generalized Easy Crossing}}
     \end{subfigure}
     
     \begin{subfigure}[b]{0.45\textwidth}
         \centering
         \includegraphics[width=\textwidth]{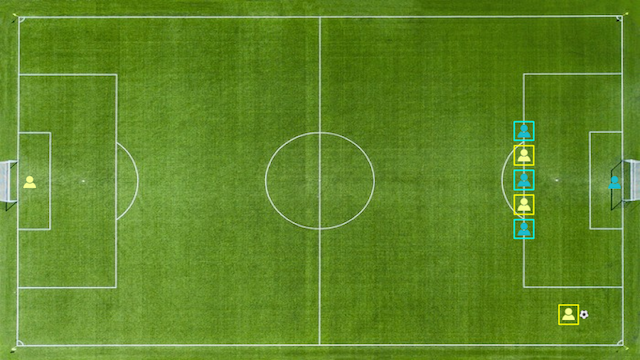}
         \caption{{\sc Hard Crossing}}
         \label{fig:ds_off_cross_hard}
     \end{subfigure}
     \hfill
     \begin{subfigure}[b]{0.45\textwidth}
         \centering
         \includegraphics[width=\textwidth]{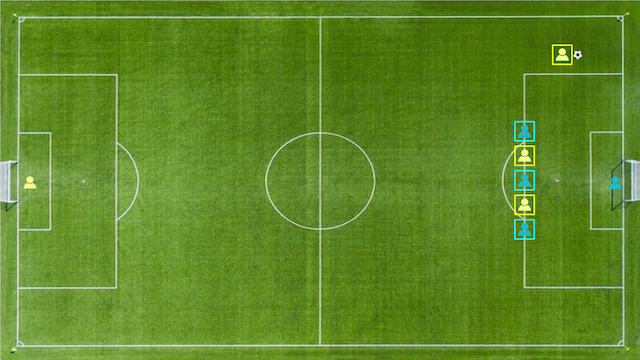}
         \caption{{\sc Generalized Hard Crossing}}
     \end{subfigure}
     
     \begin{subfigure}[b]{0.45\textwidth}
         \centering
         \includegraphics[width=\textwidth]{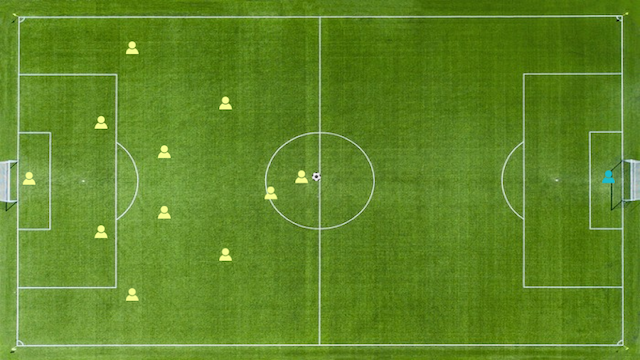}
         \caption{{\sc 11 vs GK}}
     \end{subfigure}
     \hfill
     \begin{subfigure}[b]{0.45\textwidth}
         \centering
         \includegraphics[width=\textwidth]{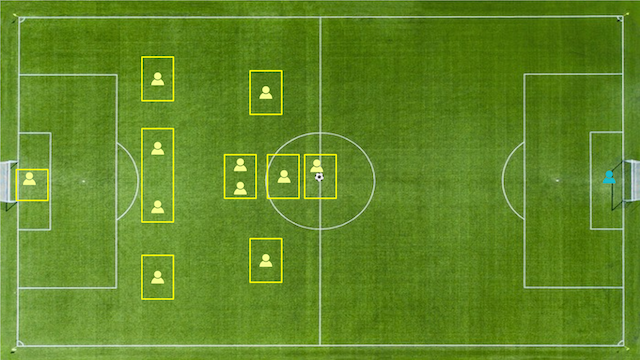}
         \caption{{\sc Generalized 11 vs GK}}
     \end{subfigure}
     
     \begin{subfigure}[b]{0.45\textwidth}
         \centering
         \includegraphics[width=\textwidth]{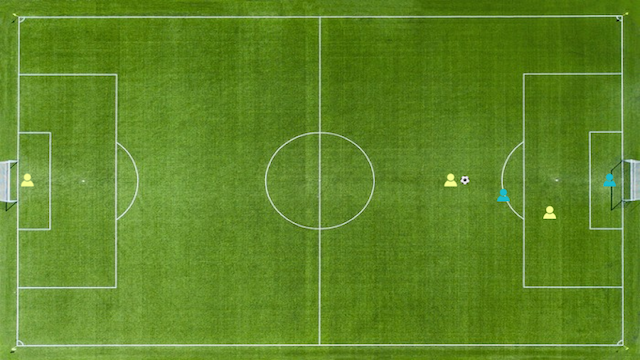}
         \caption{{\sc Avoid, pass, and Shoot}}
     \end{subfigure}
     \hfill
     \begin{subfigure}[b]{0.45\textwidth}
         \centering
         \includegraphics[width=\textwidth]{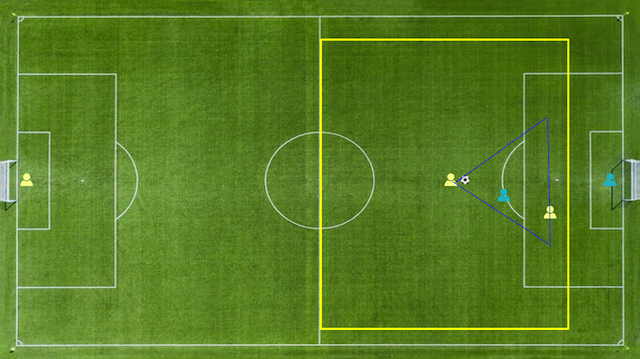}
         \caption{{\sc Generalized Avoid, pass, and Shoot}}
     \end{subfigure}

    \caption{New offense benchmark scenarios (left images) and corresponding generalized test scenarios (right images) in our dataset. The highlighted boxes represent the regions over which players' initial positions are uniformly randomly distributed. The opponent is in blue and the RL team in yellow.}
    \label{fig:scenarios}
\end{figure*}

\begin{figure*}
     \centering
     \begin{subfigure}[b]{0.45\textwidth}
         \centering
         \includegraphics[width=\textwidth]{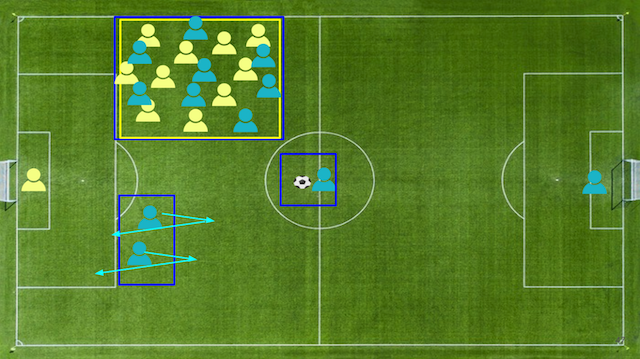}
         \caption{{\sc 11 vs 11 with Open Players}}
     \end{subfigure}
     \hfill
     \begin{subfigure}[b]{0.45\textwidth}
         \centering
         \includegraphics[width=\textwidth]{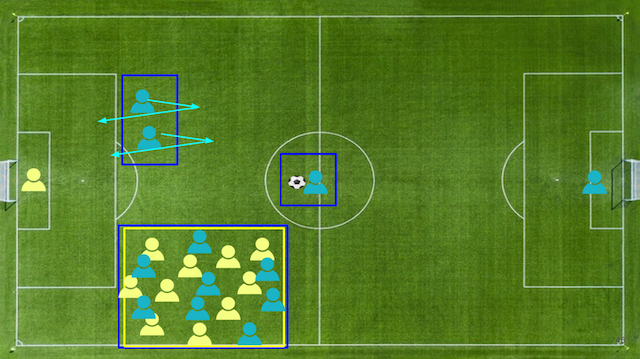}2
         \caption{{\sc Generalized 11 vs 11 with Open Players}}
     \end{subfigure}

    \caption{New offense benchmark scenarios (left images) and corresponding generalized test scenarios (right images) in our dataset. The highlighted boxes represent the regions over which players' initial positions are uniformly randomly distributed. The opponent is in blue and the RL team in yellow.}
    \label{fig:scenarios2}
\end{figure*}

In all of our six offense scenarios as shown in Fig.~\ref{fig:scenarios} and \ref{fig:scenarios2}, we explicitly modelled the initial state distribution in using \scenic and implicitly specified the behaviors to environment players by assigning the rule-based AI bots provided by Google Research Football (GRF) to control all non-RL players.

\textbf{Hard Crossing}: A very common scenario in real soccer games: 2 of our players along are guarded by 3 of the opponent players, in an interleaved manner, along the line of the penalty box. Another of our player at the edge of the field is attempting a cross. 

\textbf{11 vs GK}: Our team, with a full lineup of eleven players in a traditional 4-4-2 formation, needs to score against the opponent goalkeeper. 

\textbf{Avoid, Pass, and Shoot}: Two of our players, one starting on the middle of the right half and the other inside the penalty box, tries to score. One opponent defender starts between our players to intercept direct pass. 

\textbf{Easy Crossing}: An easy crossing scenario involving two of our players against opponent defender and goalkeeper in the penalty box. 

\textbf{11 vs 11 with Open Players}: A full game scenario where there are two unmarked players near the opponent goal. This is to test how wide "vision" an RL agent has in identifying unmarked players near the oppnent goal.

\subsubsection{Defense Scenarios} 
\begin{figure*}
     \centering
     \begin{subfigure}[b]{0.45\textwidth}
         \centering
         \includegraphics[width=\textwidth]{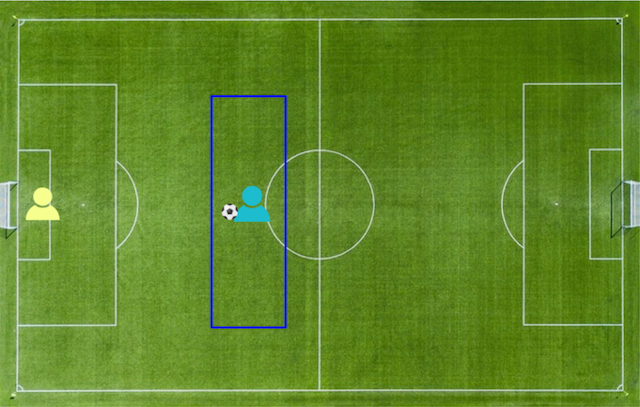}
         \caption{{\sc Goalkeeper vs Opponent}}
     \end{subfigure}
     \hfill
     \begin{subfigure}[b]{0.45\textwidth}
         \centering
         \includegraphics[width=\textwidth]{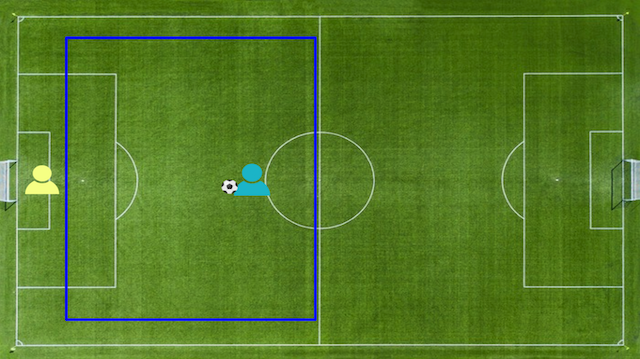}
         \caption{{\sc Generalized Goalkeeper vs Opponent}}
     \end{subfigure}
     
     \begin{subfigure}[b]{0.45\textwidth}
         \centering
         \includegraphics[width=\textwidth]{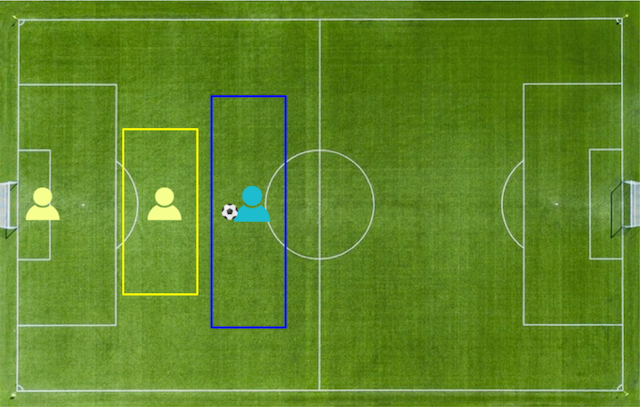}
         \caption{{\sc Defender vs Opponent with Hesitant Dribble}}
     \end{subfigure}
     \hfill
     \begin{subfigure}[b]{0.45\textwidth}
         \centering
         \includegraphics[width=\textwidth]{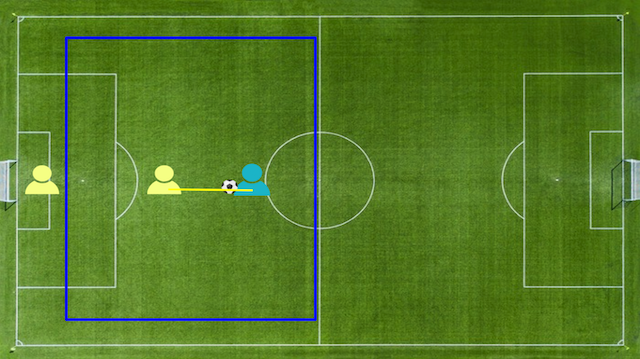}
         \caption{{\sc Generalized Defender vs Opponent with Hesitant Dribble}}
     \end{subfigure}
     
     \begin{subfigure}[b]{0.45\textwidth}
         \centering
         \includegraphics[width=\textwidth]{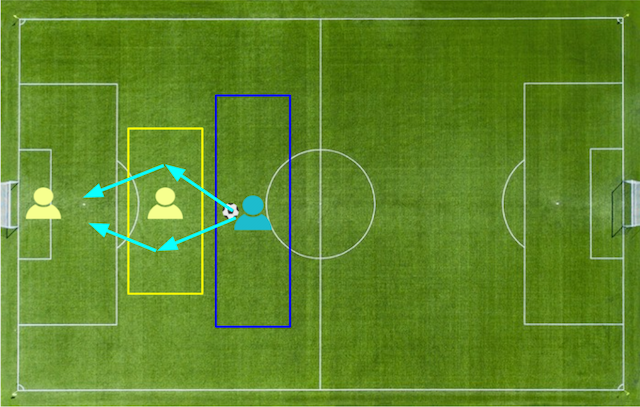}
         \caption{{\sc Defender vs Opponent with Zigzag Dribble}}
     \end{subfigure}
     \hfill
     \begin{subfigure}[b]{0.45\textwidth}
         \centering
         \includegraphics[width=\textwidth]{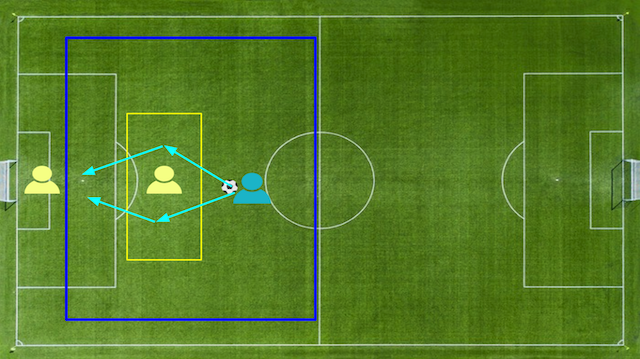}
         \caption{{\sc Generalized Defender vs Opponent with Zigzag Dribble}}
     \end{subfigure}
     
     \begin{subfigure}[b]{0.45\textwidth}
         \centering
         \includegraphics[width=\textwidth]{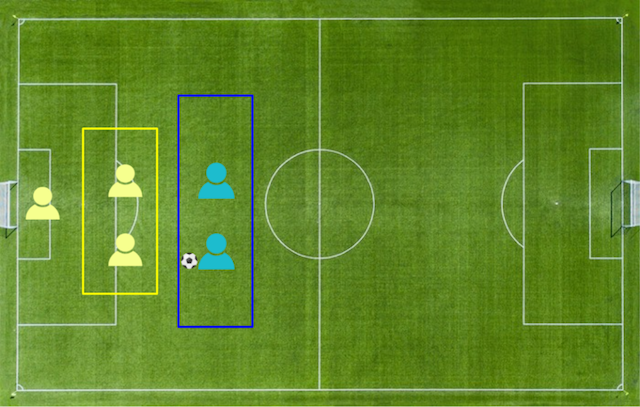}
         \caption{{\sc 2 vs 2}}
     \end{subfigure}
     \hfill
     \begin{subfigure}[b]{0.45\textwidth}
         \centering
         \includegraphics[width=\textwidth]{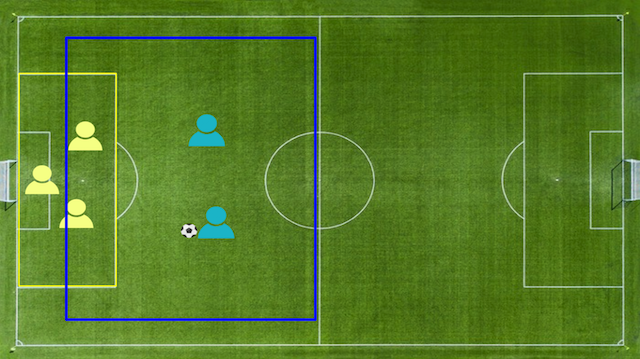}
         \caption{{\sc Generalized 2 vs 2}}
     \end{subfigure}
     
    \caption{New defense benchmark scenarios (left images) and corresponding generalized test scenarios (right images) in our dataset. The highlighted boxes represent the regions over which players' initial positions are uniformly randomly distributed. The opponent is in blue and the RL team in yellow.}
    \label{fig:defense_scenarios1}
\end{figure*}

\begin{figure*}
    \centering
    \begin{subfigure}[b]{0.45\textwidth}
         \centering
         \includegraphics[width=\textwidth]{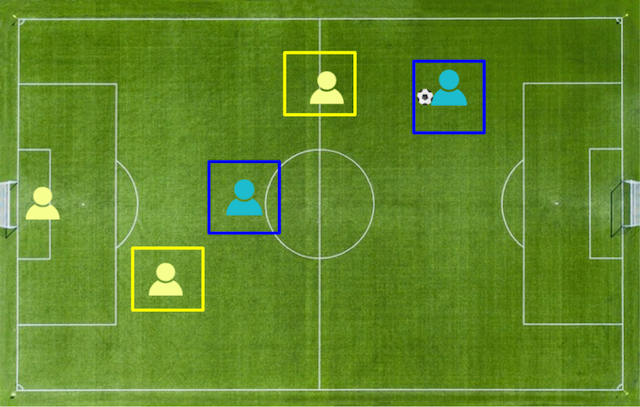}
         \caption{{\sc 2 vs 2 Counterattack}}
     \end{subfigure}
     \hfill
     \begin{subfigure}[b]{0.45\textwidth}
         \centering
         \includegraphics[width=\textwidth]{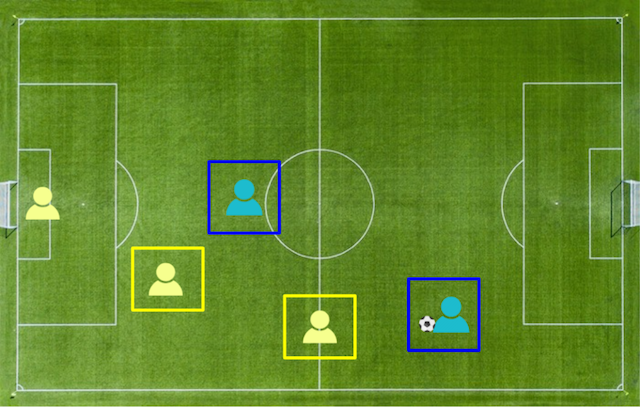}
         \caption{{\sc Generalized 2 vs 2 Counterattack}}
     \end{subfigure}
     
     \begin{subfigure}[b]{0.45\textwidth}
         \centering
         \includegraphics[width=\textwidth]{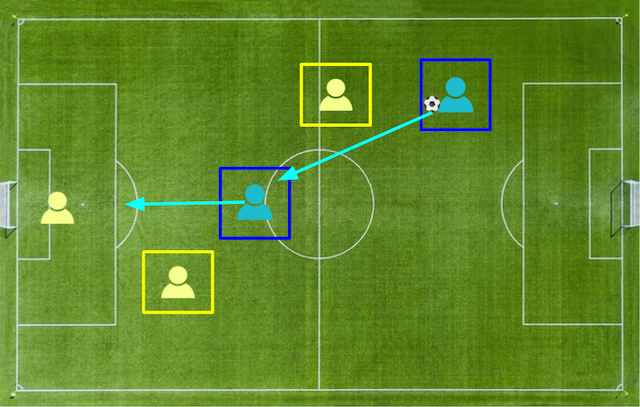}
         \caption{{\sc 2 vs 2 with High Pass Forward}}
     \end{subfigure}
     \hfill
     \begin{subfigure}[b]{0.45\textwidth}
         \centering
         \includegraphics[width=\textwidth]{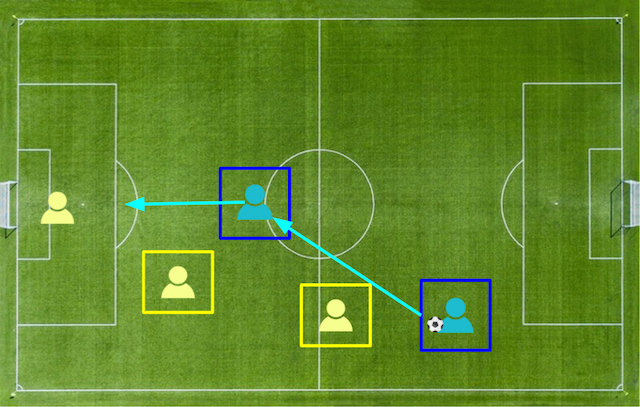}
         \caption{{\sc Generalized 2 vs 2 with High Pass Forward}}
     \end{subfigure}
     
    \begin{subfigure}[b]{0.45\textwidth}
         \centering
         \includegraphics[width=\textwidth]{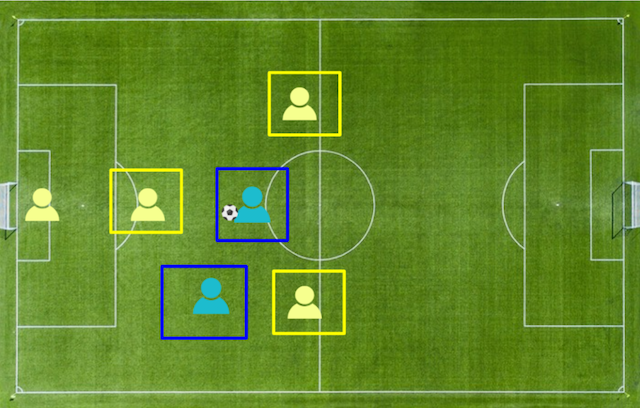}
         \caption{{\sc 3 vs 2 Counterattack}}
     \end{subfigure}
     \hfill
     \begin{subfigure}[b]{0.45\textwidth}
         \centering
         \includegraphics[width=\textwidth]{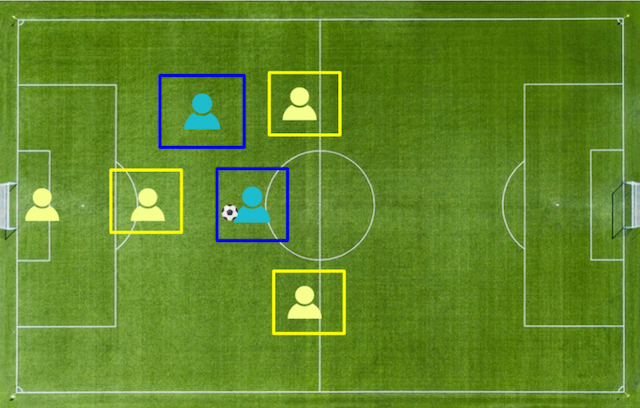}
         \caption{{\sc Generalized 3 vs 2 Counterattack}}
     \end{subfigure}
     
    \begin{subfigure}[b]{0.45\textwidth}
         \centering
         \includegraphics[width=\textwidth]{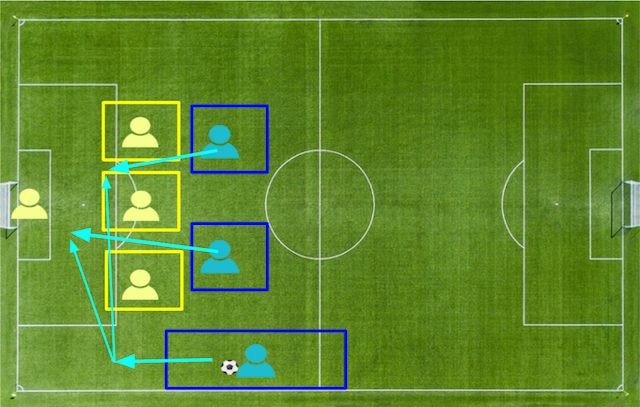}
         \caption{{\sc 3 vs 3 Cross from side}}
     \end{subfigure}
     \hfill
     \begin{subfigure}[b]{0.45\textwidth}
         \centering
         \includegraphics[width=\textwidth]{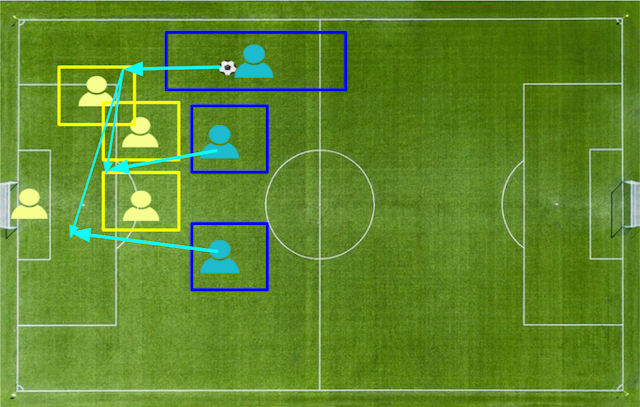}
         \caption{{\sc Generalized 3 vs 3 Cross from side}}
     \end{subfigure}
     
    \caption{New defense benchmark scenarios (left images) and corresponding generalized test scenarios (right images) in our dataset. The highlighted boxes represent the regions over which players' initial positions are uniformly randomly distributed. The opponent is in blue and the RL team in yellow.}
    \label{fig:defense_scenarios2}
\end{figure*}

Like the offense scenarios, we assigned rule-based AI bots provided by GRF by default to control non-RL players in many of our defense scenarios as shown in Fig.~\ref{fig:defense_scenarios1}, \ref{fig:defense_scenarios2}, \ref{fig:defense_scenarios3}. However, if the AI bots do not exhibit expected behavior for our modelled scenarios, we specified non-RL players' behaviors in \scenic{}. For these scenarios with specified behaviors in \scenic{}, their behaviors are highlighted with light blue arrows. 

\textbf{Goalkeeper vs Opponent}: 
This scenario is designed to train an RL agent to be a defensive goalkeeper when it has to face an opponent one-on-one. 

\textbf{Defender vs Opponent with Hesitant Dribble}:
The opponent dribbles, stop, then dribbles again in a repeated manner.

\textbf{Defender vs Opponent with Zigzag Dribble}:
This opponent aggressively evades the defender with zigzag dribble towards the goal and shoots. 

\textbf{2 vs 2}:
Typical, 2 vs 2 setting where two defenders are already in place to fend off the two opponents near the penalty area with the ball. 

\textbf{2 vs 2 Counterattack}:
An opponent attacking midfielder is already advanced deep into the left side of the field. The opponent right midfielder behind either short passes the ball to the attacking midfielder or dribbles up the field.

\textbf{2 vs 2 High Pass Forward}:
An opponent attacking midfielder is already advanced deep into the left side of the field. The opponent right midfielder quickly advances the ball to the attacking midfielder via high pass. 

\textbf{3 vs 2 Counterattack}:
The defender near the penalty box is temporarily outnumbered by the opponent players due to a sudden counterattack. 

\textbf{3 vs 3 Cross from Side}:
The opponent player on the side crosses the ball to either of the teammates in the middle who are running towards the penalty box to receive the ball. \footnote{The {\sc Defense 3vs3 with cross} scenario doesn't conclude a game upon  a change in ball possession, unlike other scenarios}

\textbf{3 vs 3 Side Build Up Play}:
Instead of crossing, the opponent player on the side builds up a play by short passing to its teammates.

\subsection{Testing Generalization}

\begin{figure*}
    \centering
    \begin{subfigure}[b]{0.45\textwidth}
         \centering
         \includegraphics[width=\textwidth]{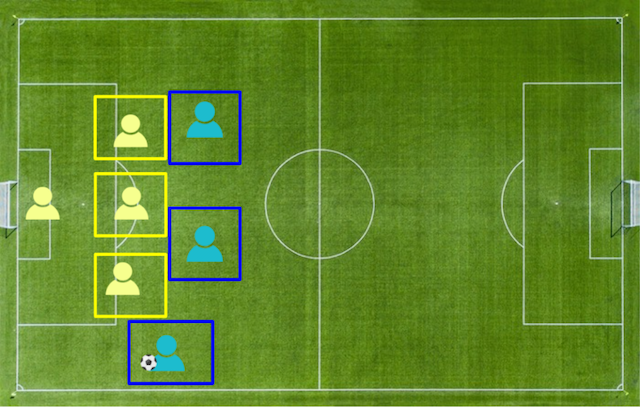}
         \caption{{\sc 3 vs 3 side build up play}}
     \end{subfigure}
     \hfill
     \begin{subfigure}[b]{0.45\textwidth}
         \centering
         \includegraphics[width=\textwidth]{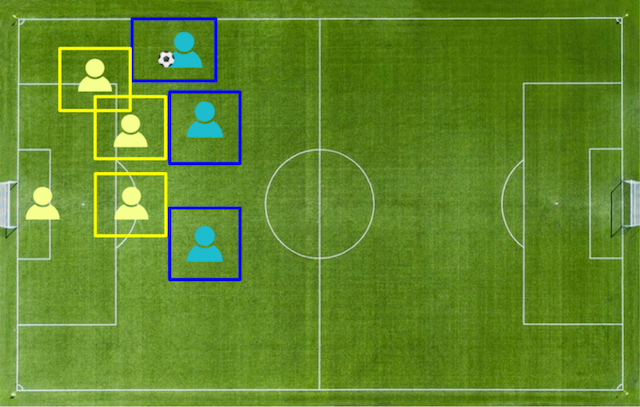}
         \caption{{\sc Generalized 3 vs 3 side build up play}}
     \end{subfigure}
    \caption{New defense benchmark scenario (left images) and corresponding generalized test scenarios (right images) in our dataset. The highlighted boxes represent the regions over which players' initial positions are uniformly randomly distributed. The opponent is in blue and the RL team in yellow.}
    \label{fig:defense_scenarios3}
\end{figure*}

For all the new benchmark scenarios in our dataset as well as for the selected five GRF's scenarios, we generalized those scenarios to test the generalizability of the trained RL agent. Our test scenarios are juxtaposed to corresponding scenarios in Fig.~\ref{fig:scenarios}, ~\ref{fig:defense_scenarios1}, ~\ref{fig:defense_scenarios2}, ~\ref{fig:defense_scenarios3}. We modelled these test scenarios by either (i) adding distribution over the initial state or (ii) creating a symmetric opposite formation. 

\subsection{Semi-Expert Stochastic Policies}
\begin{figure*}
    \centering
    \begin{subfigure}[b]{0.45\textwidth}
         \centering
         \includegraphics[width=\textwidth]{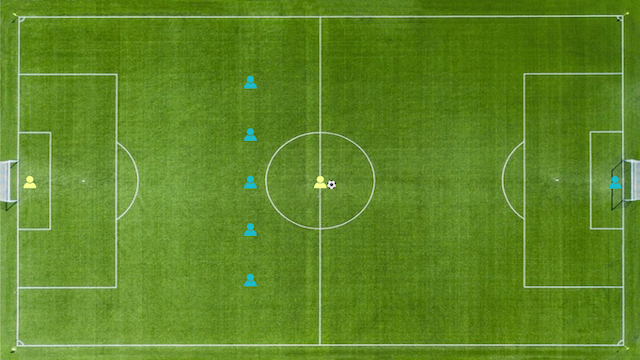}
         \caption{{\sc run to score with a goal keeper}}
     \end{subfigure}
     \hfill
     \begin{subfigure}[b]{0.45\textwidth}
         \centering
         \includegraphics[width=\textwidth]{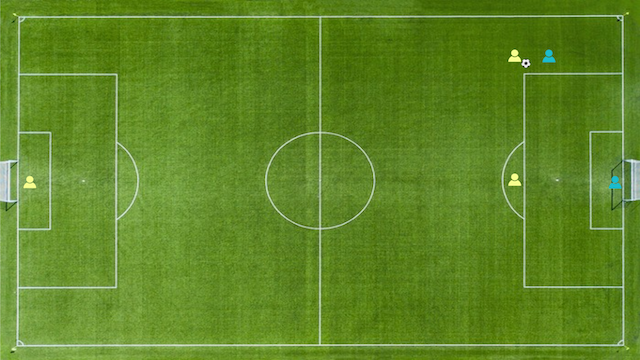}
         \caption{{\sc Pass and Shoot with a Goal Keeper}}
     \end{subfigure}
     
     \begin{subfigure}[b]{0.45\textwidth}
         \centering
         \includegraphics[width=\textwidth]{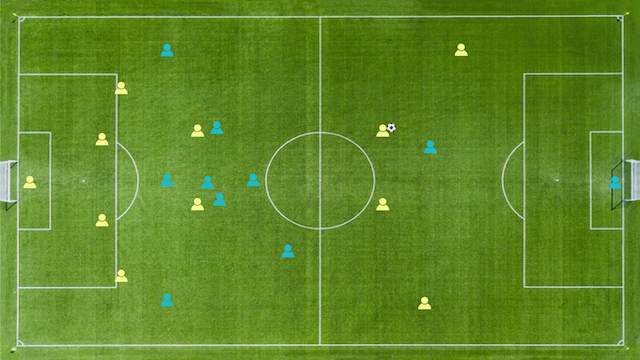}
         \caption{{\sc easy counterattack}}
     \end{subfigure}
    \caption{Google Research Football environment's scenarios for which we wrote semi-expert RL policies}
    \label{fig:grf_scenarios}
\end{figure*}

We selected five scenarios from GRF's and our benchmark scenarios. The selected GRF's scenarios are shown in Fig.~\ref{fig:grf_scenarios}. The following are the brief descriptions of policies encoded in \scenic{} for each scenario. 

\textbf{Pass and Shoot with a Goal Keeper}:
We randomly choose one of the two policies for the RL agent. In the first policy, the player dribbles to the penalty area and shoots once inside it. In the second policy, the player passes the ball to the teammate, who will then dribble towards the goal and shoot.

\textbf{Easy Counterattack}:
The first player will pass the ball to the right midfielder. Then, the player with the ball will run into the penalty area, and if there is an opponent player on the way, the player will pass the ball to the nearest teammate. The player will shoot at a corner of the goal once inside the penalty area.

\textbf{Run to Score with a Goal Keeper}:
The player with the ball will first sprint towards the goal and turn slightly to either left or right randomly to evade the opponent goalkeeper’s interception. Once the player bypasses the goalkeeper, or is inside the penalty area, the player will shoot.

\textbf{Avoid Pass and Shoot}: 
The player decides to go towards 3 suitable regions of scoring: left edge/ middle/ right edge of the right goal post, by keeping as much distance possible to the opponent defender. At each time step it decides one of the three destination location. It first predicts its next position for all the three suitable destinations and pick the direction which keeps it farthest of the opponent. If the player comes near the defender it passes the ball to its teammate and if it can successfully go near the goal post, it attempts shooting.

\textbf{11 vs Goal Keeper}:
The player with the ball runs towards the right goalpost, if it reaches near the goal post it attempts to shoot. If the opponent goal keeper comes near (i.e. within seven meters) our player before it can reach near the right goal post, it stops running and shoots immediately.

\section{On Our \scenic{} Libraries}
Users can quickly model scenarios by referencing models, actions, and behaviors from the libraries that we open-sourced along with our interface. To see our library codes, please refer to our attached README pdf file for the pathways to these libraries. 

\subsection{Model Library}
The model library defines three categories of objects. First, it defines different regions of the field such as penalty box area. Second, it defines the \texttt{Ball}. Lastly, it defines the \texttt{Player}. There are two types of player objects which inherit this \texttt{class Player}: Left and Right players. The left players represent the RL team, and the right, the opponent. Within each team, the players are further classified into different roles. The naming convention is ``the team + role abbreviated in two letters." For example, the left team's goalkeeper is defined as texttt{LeftGK}. Likewise, for the right team players. 

\subsection{Action Library}
This library defines the action space of any players. It consists of twelve different actions such as \texttt{Pass}, \texttt{Shoot}, \texttt{Dribble}, \texttt{Sprint}, \texttt{Slide}, etc. These actions can be referenced in the \scenic{} script using the syntax, \textit{take}. For example, to take sliding action, users can write \texttt{take Slide()} in their programs. 

\subsection{Behavior Library}
The behavior library consists of basic soccer skills that we modelled in \scenic{}. This library consists of helper functions defined using the syntax, \texttt{def}, and behaviors, which reference those helper functions, are defined with the syntax, \texttt{behavior}. For brevity, we refer the reviewers to our annotated library code.

\section{Details on Experimental Setup and Training}
We use the OpenAI Baselines'~\cite{baselines} implementation of PPO. The training was run for 5M timesteps with 16 parallel workers. All of our experiments are run on g4dn.4xlarge instances on Amazon AWS: a machine with a single NVIDIA T4 gpu, 16 virtual cores and 64GB RAM.

{\bf Network architecture \& Hyperparameters}
For the PPO training, we first experimented with the network architecture and hyperparameters from~\cite{kurach2020google} and was able to reproduce their result. \cite{kurach2020google} did an extensive  search to select their hyperparameters and hence we decided to use the same for our experiments. The architecture we used from ~\cite{kurach2020google} is similar to the architecture introduced in~\cite{impala}, with the exception of using four big blocks instead of three.

Table~\ref{tab:hp_ppo} provides specific values of the hyperparameters used in the PPO experiments. 

\begin{table}[!h]
\centering
\begin{tabular}{lr}
\hline
\textbf{Parameter}           & \textbf{Value}   \\
\hline
Action Repetitions  & 1       \\
Clipping Range          & .115    \\
Discount Factor ($\gamma$)     & 0.997   \\
Entropy Coefficient & 0.00155 \\
GAE ($\lambda$)                 & 0.95    \\
Gradient Norm Clipping          & 0.76  \\ 
Learning Rate & 0.00011879\\
Number of Actors & 16\\
Optimizer & Adam \\
Training Epochs per Update & 2 \\
Training Min-batches per Update & 4 \\
Unroll Length/n-step & 512 \\
Value Function Coefficient & 0.5 \\ 
\hline 
\end{tabular}
\caption{Training Parameters for PPO.}\label{tab:hp_ppo}
\end{table}

The parameters for behavior cloning is shown in Table~\ref{tab:hp_bc}. For the GRF academy scenarios the behavior cloning algorithm is run for 16 epochs while for the offense scenarios it was run for 5 epochs. 

\begin{table}[!h]
\centering
\begin{tabular}{ll}
\hline
\textbf{Parameter}           & \textbf{Value}   \\
\hline
Learning Rate & 3e-4\\
Batch Size & 256\\
Optimizer & Adam \\
Epsilon(Adam) & 1e-5\\
\hline 
\end{tabular}
\caption{Training Parameters for Imitation Learning.}\label{tab:hp_bc}
\end{table}

\section{Interface details and Reproducibility}

Our interface follows the widely used OpenAI Gym API~\cite{brockman2016openai}.  For sample usage, we refer readers to our code that is submitted along with this supplement. The code contains necessary scripts, and the attached README pdf file contains detailed description of all our API with examples and a link to a google drive which contains all our trained checkpoints and training logs. 

\section{Performance}

As we are adding an additional layer over the GRF simulator, we wanted to measure how much overhead we are adding over the base GRF simulator. We selected five GRF academy scenarios and ran a simulation of 20K timesteps with a random policy both in the GRF simulator and in our interface. The simulation was ran sequentially, i.e., no parallelism was used. Across the scenarios, the GRF simulator took an average of 74.28 seconds for executing a simulation of 20K timesteps, while our interface took 222.07 seconds, showing a 2.99x drop in speed. Some of this overhead is inevitable however, we believe there are ways to speed up . First, as we change the initial state every episode/simulation: we update the Python scenario file used by the GRF simulator for each episode/simulation. We plan to modify GRF interface to avoid such disk-access each simulation to speed up among other performance improvements. The scenarios we used for the experiments are namely: i) Empty Goal, ii) Empty Goal Close, iii) Pass and Shoot with Keeper, iv) Run, Pass, and Shoot with Keeper, and v) Run to Score with Keeper.


%% file: Supplimentary Materials/supplementary.tex
\textbf{Supplementary Materials}

\section{Assets & Accessability} 

Our dataset and code for reproduction can be found in: \href{https://sites.google.com/view/scenic4rl/}{https://sites.google.com/view/scenic4rl/} 

\wl{Eddie}

\section{Dataset Documentation and Intended Uses}
\wl{Eddie} 
Comment about Dataset Documentation, Website, Maintenance Plan, Licensing, Reproduction

\newlist{} 
Submission introducing new datasets must include the following in the supplementary materials:

1. Dataset documentation and intended uses. Recommended documentation frameworks include datasheets for datasets, dataset nutrition labels, data statements for NLP, and accountability frameworks. 

2. URL to website/platform where the dataset/benchmark can be viewed and downloaded by the reviewers.

3. Author statement that they bear all responsibility in case of violation of rights, etc., and confirmation of the data license.

4. Hosting, licensing, and maintenance plan. The choice of hosting platform is yours, as long as you ensure access to the data (possibly through a curated interface) and will provide the necessary maintenance.

--- For benchmarks, the supplementary materials must ensure that all results are easily reproducible. Where possible, use a reproducibility framework such as the ML reproducibility checklist, or otherwise guarantee that all results can be easily reproduced, i.e. all necessary datasets, code, and evaluation procedures must be accessible and documented.

\section{Dataset Documentation: Description of proposed Scenarios and Policies}
~\label{sec:dataset_descrition}
In this section we briefly describe each of our proposed scenarios. 

\begin{figure}
     \centering
     \begin{subfigure}[b]{0.45\textwidth}
         \centering
         \includegraphics[width=\textwidth]{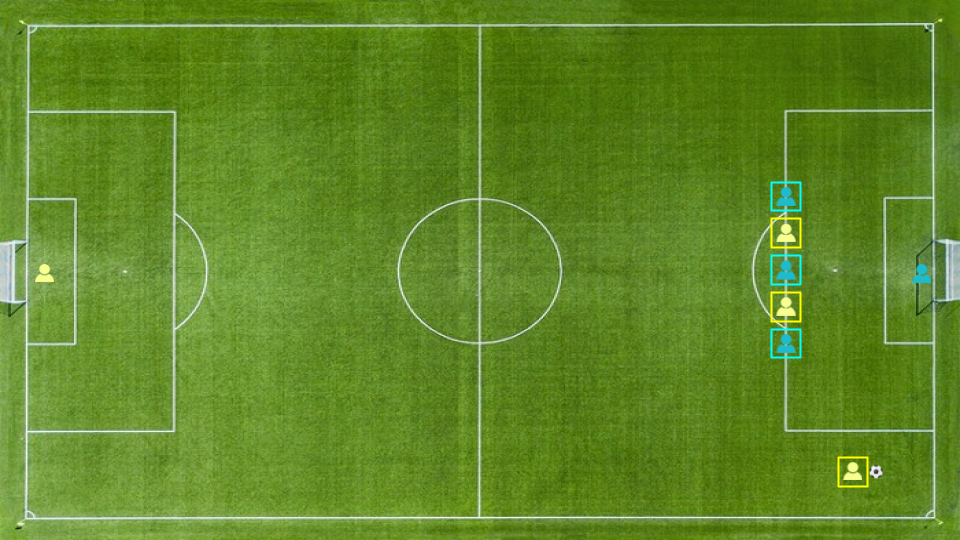}
         \caption{{\sc Hard Crossing}}
         \label{fig:ds_off_cross_hard}
     \end{subfigure}
     \hfill
     \begin{subfigure}[b]{0.45\textwidth}
         \centering
         \includegraphics[width=\textwidth]{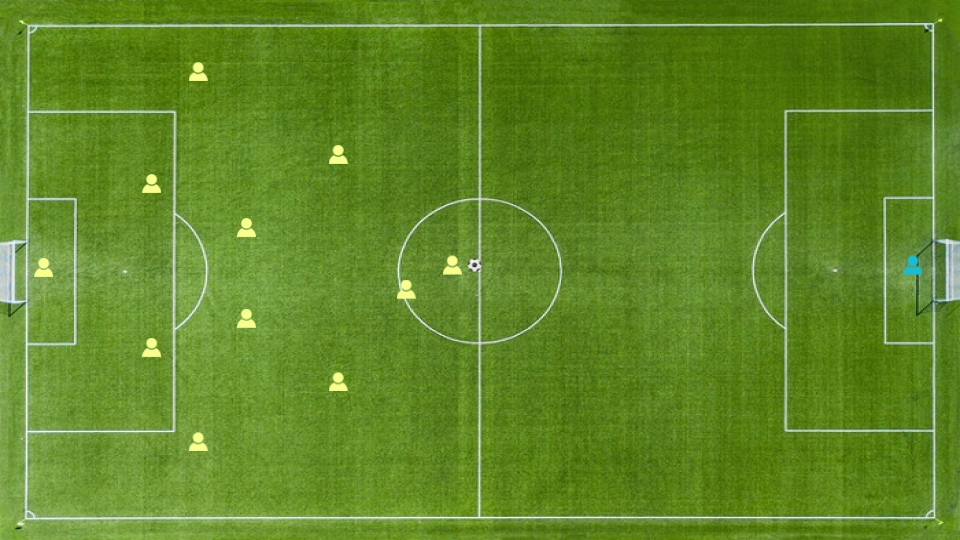}
         \caption{{\sc 11 vs GK}}
         \label{fig:ds_off_11v1}
     \end{subfigure}
     
     \begin{subfigure}[b]{0.45\textwidth}
         \centering
         \includegraphics[width=\textwidth]{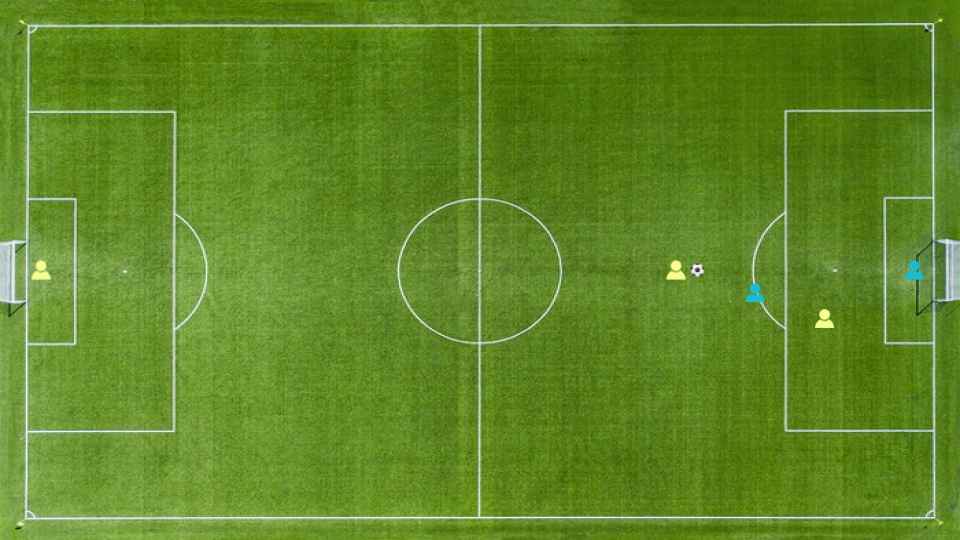}
         \caption{{\sc Avoid, pass, and Shoot}}
         \label{fig:ds_off_avoid_pass_shoot}
     \end{subfigure}
     \hfill
     \begin{subfigure}[b]{0.45\textwidth}
         \centering
         \includegraphics[width=\textwidth]{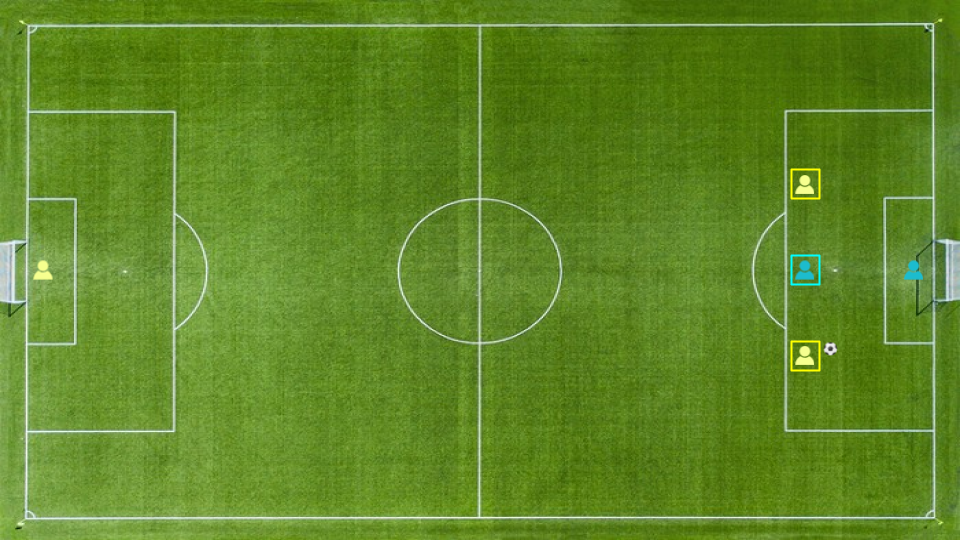}
         \caption{{\sc Easy Crossing}}
         \label{fig:ds_off_cross_easy}
     \end{subfigure}
    \caption{New Scenarios proposed in this paper.}
        \label{fig:test_new_offense}
\end{figure}

\textbf{{\sc Offense Scenarios}}

\textbf{Hard Crossing}: A very common scenario in real soccer games: 2 of our players along are guarded by 3 of the opponent players, in an interleaved manner, along the line of the penalty box. Another of our player at the edge of the field is attempting a cross. 

\textbf{11 vs GK}: Our team, with a full lineup of eleven players in a traditional 4-4-2 formation, needs to score against the opponent goalkeeper. 

\textbf{Avoid, Pass, and Shoot}: Two of our players, one starting on the middle of the right half and the other inside the penalty box, tries to score. One opponent defender starts between our players to intercept direct pass. 

\textbf{Easy Crossing}: An easy crossing scenario involving two of our players against opponent defender and goalkeeper in the penalty box. 

\textbf{{\sc Defensive Scenarios}}

\azad{To be filled up by Eddie}

\section{Details on Experimental Setup and Reproducability}
We use the OpenAI Baselines'~\cite{baselines} implementation of PPO. The training was run for 5M timesteps with 16 parallel workers. All of our experiments are run on g4dn.4xlarge instances on Amazon AWS: a machine with a single NVIDIA T4 gpu, 16 virtual cores and 64GB RAM.

\subsection{Network architecture \& Hyperparameters}
For the PPO training, we first experimented with the hyperparameters from~\cite{kurach2020google} and was able to reproduce their result. \cite{kurach2020google} did an extensive hyperparameter search to select their hyperparameters and hence we decided to use the same for our experiments. Table~\ref{tab:hp_ppo} provides specific values of the hyperparameters.

\begin{table}[!h]
\begin{tabular}{lr}
\hline
\textbf{Parameter}           & \textbf{Value}   \\
\hline
Action Repetitions  & 1       \\
Clipping Range          & .115    \\
Discount Factor ($\gamma$)     & 0.997   \\
Entropy Coefficient & 0.00155 \\
GAE ($\lambda$)                 & 0.95    \\
Gradient Norm Clipping          & 0.76  \\ 
Learning Rate & 0.00011879\\
Number of Actors & 16\\
Optimizer & Adam \\
Training Epochs per Update & 2 \\
Training Min-batches per Update & 4 \\
Unroll Length/n-step & 512 \\
Value Function Coefficient & 0.5 \\ 
\hline 
\end{tabular}
\caption{Training Parameters for PPO.}\label{tab:hp_ppo}
\end{table}

The parameters for behavior cloning is shown in Table~\ref{tab:hp_bc}

\begin{table}[!h]
\begin{tabular}{ll}
\hline
\textbf{Parameter}           & \textbf{Value}   \\
\hline
Learning Rate & 3e-4\\
Batch Size & 256\\
Optimizer & Adam \\
Epsilon(Adam) & 1e-5\\
\hline 
\end{tabular}
\caption{Training Parameters for Imitation Learning.}\label{tab:hp_bc}
\end{table}

\section{Interface details }
\subsection{Scenic GFootball Interface: API \& Sample Usage}

Our Platform follows the widely used OpenAI Gym API~\cite{brockman2016openai}.  Below we show an example code of how to instantiate an environment from a scenario file along with other parameters in a dictionary and run a random agent on it. 

\begin{verbatim}
from scenic.simulators.gfootball.rl.gfScenicEnv import GFScenicEnv
scenario = f"path/script.scenic"
env = GFScenicEnv(scenario=scenario, allow_render=False, 
                  gf_env_settings={"rewards":'scoring', ...})
env.reset() 
done = False 
while not done:
    action = env.action_space.sample() 
    observation, reward, done, info = env.step(action)
\end{verbatim}

For detailed usage we refer our readers to our website. 

\azad{add details on 1) how to enable/scenic behavior, 2) generate demonstration data ... 3) read demonstration data? here or in url ??}